\newif\ifusecompsoc
\usecompsocfalse 
\usecompsoctrue  

\ifusecompsoc
    \documentclass[10pt,journal,compsoc]{IEEEtran}
\else
    \documentclass[lettersize,journal]{IEEEtran}
\fi

\usepackage{amsmath,amsfonts}
\usepackage{algorithm}
\usepackage{algorithmic}
\usepackage{array}
\usepackage[caption=false,font=normalsize,labelfont=sf,textfont=sf]{subfig}
\usepackage{textcomp}
\usepackage{stfloats}
\usepackage{url}
\usepackage{verbatim}
\usepackage{graphicx}
\usepackage{cite}
\usepackage{amsmath, amsfonts, amssymb, bm}
\usepackage{booktabs}
\usepackage{multirow}
\usepackage{adjustbox}
\usepackage{threeparttable}
\usepackage[mathscr]{eucal} 

\usepackage{tabularx}
\usepackage{colortbl}
\newcolumntype{Y}{>{\centering\arraybackslash}X}

\usepackage{xcolor}
\usepackage{pifont}
\usepackage{enumitem}   
\usepackage[most]{tcolorbox} 
\tcbuselibrary{breakable} 

\usepackage{tikz}
\usetikzlibrary{tikzmark, calc, matrix, positioning}
\tikzset{
    solidbox/.style={draw=black, thick, solid, line width=0.5pt},
    dashedbox/.style={draw=black, thick, dashed, line width=0.5pt}
}

\definecolor{royalblue}{HTML}{F6F8FD}
\definecolor{royalpink}{HTML}{FCF8FD}
\definecolor{tablegray}{gray}{.95}
\definecolor{tableblue}{rgb}{0.957, 0.962, 0.992} 

\tcbset{
    royalbluebox/.style={
        enhanced, 
        breakable=true, 
        colback=royalblue, 
        colframe=white!10,
        boxrule=1pt, 
        left=0.8mm, right=0.8mm, 
        top=0.5mm, bottom=0.5mm, 
        before skip=8pt, 
        after skip=8pt,
        pad at break=5pt, 
    }
}
\tcbset{
    royalpinkbox/.style={
        enhanced, 
        breakable=true, 
        colback=royalpink, 
        colframe=white!10, 
        boxrule=1pt,
        left=0.5mm, right=0.5mm,
        top=0.3mm, bottom=0.3mm, 
        before skip=8pt,
        after skip=8pt,
        pad at break=5pt,
    }
}
\tcbset{
    blacklinebox/.style={
        enhanced, 
        breakable=true, 
        colback=white, 
        colframe=black!80, 
        boxrule=0.5pt, 
        left=0.5mm, right=0.5mm,
        top=0.3mm, bottom=0.3mm, 
        before skip=8pt, 
        after skip=8pt, 
        pad at break=5pt, 
    }
}

\usepackage[amsmath,thmmarks,framed]{ntheorem} %
\theoremstyle{plain}
\usepackage{framed}
\usepackage{color}
\definecolor{ora}{RGB}{245, 180, 130}
\definecolor{gray}{RGB}{160, 160, 160}
\newframedtheorem{define}{Definition}
\newtheorem{assumption}{Assumption}
\newtheorem{theorem}{Theorem}
\newtheorem{remark}{Remark}

\newcommand{\cmark}{\textcolor{green}{\ding{51}}} %
\newcommand{\xmark}{\textcolor{red}{\ding{55}}}   %

\renewcommand{\algorithmiccomment}[1]{\quad \textcolor{gray}{\textit{\# #1}}}

\begin{document}

\title{ 
FastBUS: A Fast Bayesian  Framework for Unified Weakly-Supervised Learning
}

\author{Ziquan Wang,  Haobo Wang, Ke Chen, Lei Feng,  Gang Chen 
\IEEEcompsocitemizethanks{
\IEEEcompsocthanksitem Ziquan Wang, and Haobo Wang are with the School of Software Technology, Zhejiang University, Hangzhou 310027, China. Email: \{ziquanwang, wanghaobo\}@zju.edu.cn 
\IEEEcompsocthanksitem Ke Chen are with the State Key Laboratory of Blockchain and Data Security, Zhejiang University, Zhejiang 310027, China, and also with the Hangzhou High-Tech Zone (Binjiang) Institute of Blockchain and Data Security, Binjiang 310000, China. Email: chenk@zju.edu.cn
\IEEEcompsocthanksitem Lei Feng is with the School of Computer Science and Engineering, Southeast University, Nanjing 210096, China. Email: fenglei@seu.edu.cn
\IEEEcompsocthanksitem Gang Chen are with the College of Computer Science and Technology, Zhejiang University, Hangzhou 310027, China. Email: cg@zju.edu.cn 
\IEEEcompsocthanksitem Corresponding author: Haobo Wang.
}
}

\markboth{Journal of \LaTeX\ Class Files,~Vol.~14, No.~8, August~2021}%
{Wang \MakeLowercase{\textit{et al.}}: FastBUS: A Fast Bayesian  Framework for Unified Weakly-Supervised Learning}


\newcommand{\AbstractText}{
Machine Learning often involves various imprecise labels, leading to diverse weakly supervised settings. While recent methods aim for universal handling, they usually suffer from complex manual pre-work, ignore the relationships between associated labels, or are unable to batch process due to computational design flaws, resulting in long running times. To address these limitations, we propose a novel general framework that efficiently infers latent true label distributions across various weak supervisions. Our key idea is to express the label brute-force search process as a probabilistic transition of label variables, compressing diverse weakly supervised DFS tree structures into a shared Bayesian network. From this, we derived a latent probability calculation algorithm based on generalized belief propagation and proposed two joint acceleration strategies: 1) introducing a low-rank assumption to approximate the transition matrix, reducing time complexity; 2) designing an end-to-end state evolution module to learn batch-scale transition matrices, facilitating multi-category batch processing. In addition, the equivalence of our method with the EM algorithm in most scenarios is further demonstrated. Extensive experiments show that our method achieves SOTA results under most weakly supervised settings, and achieves up to hundreds of times faster acceleration in running time compared to other general methods.
}

\newcommand{\KeywordText}{
    General Weakly Supervised Learning, Bayesian Network, Generalized Belief Propagation Algorithm
}

\ifusecompsoc
    \IEEEtitleabstractindextext{%
        \begin{abstract}   
        \AbstractText  
        \end{abstract}
    }
    \maketitle
    \IEEEdisplaynontitleabstractindextext
    \IEEEpeerreviewmaketitle
\fi
\ifusecompsoc\else   
    \maketitle
    \begin{abstract}   \AbstractText  \end{abstract}
\fi

\IEEEraisesectionheading{\section{Introduction}\label{sec:introduction}}

\IEEEPARstart{N}{owadays}, an increasing number of research facts highlight the critical role of large-scale data in achieving high-performance machine learning models \cite{ScalingLaw, ScalingLaw2, DataLaw}. However, obtaining large amounts of high-quality annotated data remains challenging. This often leads to practical issues like image labels with noise or uncertainty \cite{PartialImage, NoisySeg}, incomplete annotations in sensitive data \cite{UnlabeledDetection1, UnlabeledDetection2}, and paired or group-based datasets in biological research or information retrieval systems \cite{MultiInsProtein, PairwiseRetrieval1, PairwiseRetrieval2}. These imprecise datasets have driven significant interest in weakly supervised learning.

Weakly supervised learning encompasses several subfields tailored to distinct annotation challenges. Its key directions include noisy label (mislabeled data) or partial label learning (uncertain candidate annotations) \cite{DataUncertainty}, positive unlabeled or semi-supervised learning (incomplete labels) \cite{PosUnlSurvey, SemiSurvey}, multi-instance learning (grouped instances) or pairwise supervision (pairwise instances) \cite{MultiInsProtein, paircomp2010, PairwiseSurvey2}. To solve these problems, specialized techniques like loss correct methods for noisy labels \cite{TransMatix2017}, pseudo labeling for unlabeled data \cite{UnlabeledDetection2022, UnlabeledDetection2024}, pairwise reasoning \cite{PairwiseLLMReason2025} have been created. Despite progress, developing customized algorithms for each setting remains complex and time-consuming in engineering. Therefore, some studies turn to general weakly supervised algorithms applicable to diverse settings \cite{GWSL_NIPS_2020, GWSL_ICML_2023, GWSL_TPAMI, GWSL_ICML_2024}. CEGE \cite{GWSL_TPAMI} separates specific losses for weak supervision into label-independent and -dependent parts, where the latter is estimated using pre-synthesized datasets. Count Loss \cite{GWSL_NIPS_2023} converts special loss calculation into estimations of count probability. They are all constrained by specific losses, while other methods, free from losses, decompose data generation processes into \textbf{different graph structures}, and dissect the weak-supervised generality into a new \textbf{open problem}:
\begin{tcolorbox}[blacklinebox]
    Given weak labels \(\bm{W}\), how to compute the \textbf{latent probability} \(P(\bm{Y}|\bm{X}, \bm{W})\) for unobservable instance labels \(\bm{Y}\)?
\end{tcolorbox} 
\noindent Where \(\bm{W}\) is the weak supervised annotation (weak label), \(\bm{X}\) is the input, and \(\bm{Y}\) is the instance-level label. UUM \cite{GWSL_ICML_2023} enumerates all possible instance labels \(\bm{Y}\), similar to decomposing the problem into DFS trees. GLWS \cite{GWSL_ICML_2024} decomposes different weakly supervised scenarios into customized \textit{Nondeterministic Finite Automata} (NFA), and calculates probabilities on its trellis using forward-backward algorithms.

\begin{table}[!t]
\centering
\caption{Comparison Between Several Works and FastBUS for General Weak Supervision.}
\label{tab: related work}
\begin{threeparttable}
\begin{tabular}{@{}>{\centering\arraybackslash}m{1.4cm}
                >{\centering\arraybackslash}m{1.15cm}
                >{\centering\arraybackslash}m{0.9cm}
                >{\centering\arraybackslash}m{1.5cm}
            >{\centering\arraybackslash}m{1.55cm}@{}}
\toprule
\textbf{Strategy}   & \textbf{Method}        & 
\textbf{No Pre-Work} & \textbf{Multi-label Correlation} & \textbf{Worst-Case Time Complexity}  \\
\midrule
\multirow{2}{1.5cm}{Inductive Properties of Base Loss}     
  & Centroid Loss  \cite{GWSL_TPAMI}
  &      \xmark      &    \xmark                       &    \(-\)\ \tnote{a}                      \\
\cmidrule(lr){2-5}                  
    & Count Loss   \cite{GWSL_NIPS_2023}
    &      \cmark    &    \xmark                           &   \(\mathcal{O}(K^2)\)                         \\
\cmidrule(lr){1-5}
\multirow{3}{1.5cm}{Calculation of Latent
    Probobilities}
  & \multirow{1}{*}{UUM \cite{GWSL_ICML_2023}}          
  &   \cmark     &   \xmark       
  &   \(\mathcal{O}(C^K)\)                    \\
\cmidrule(lr){2-5} 
  & \multirow{1}{*}{IMP\cite{GWSL_ICML_2024}}     
  &   \xmark      &    \xmark                        &     \(\mathcal{O}(K|W|^2)\) \tnote{b}                  \\
\cmidrule(lr){2-5}
    & \multirow{1}{*}{FastBUS~}         
    &  \cmark    &   \cmark 
    &   \(\mathcal{O}(K|W|)\)  \\  
\bottomrule    
\end{tabular}
\begin{tablenotes}
    \item{a} \ Because of the differences between frameworks, the time complexity cannot be compared.
    \item{b} \ \(K\) denotes the number of instances in one bag, and \(|W|\) presents the number of values of weak label \(W\).
\end{tablenotes}
\end{threeparttable}
\end{table}

Although general weakly supervised methods have achieved good results, there still exist drawbacks shown in Table \ref{tab: related work}: \romannumeral1)
Some algorithms require pre-work. For example, CEGE synthesizes datasets in advance to estimate the label-dependent components; GLWS requires pre-designed NFAs for the given weakly-supervised setting, which also involves hard-coding. \romannumeral2) Current methods mostly treat different categories independently, making them hard to generalize to multi-label (ML) problems where relevant labels could occur simultaneously. \romannumeral3) The time complexity of these methods remains high. For instance, UUM suffers from exponential complexity due to label enumeration, and GLWS is quadratic in terms of the number of NFA states. Additionally, in some settings, GLWS’s NFA structure depends on the weak labels \(W\); since the labels of mini-batch samples or different categories often vary, this forces per-item (instance- and category-specific) calculation without batch processing, incurring high time costs.

To cope with these aforementioned limitations, this paper proposes a novel universal framework that efficiently and consistently adapts to diverse weakly supervised scenarios. To calculate latent probability, starting from the enumeration idea in UUM \cite{GWSL_ICML_2023}, we express the brute-force search process as a probability transition process of label random variables, thereby decomposing and compressing all weakly supervised DFS tree structures into a unified Bayesian network structure. \romannumeral1) Therefore, this Bayesian network structure naturally possesses the brute force property of DFS, which not only eliminates complex pre-work (pre-designed NFAs \cite{GWSL_ICML_2024} or pre-synthesized dataset \cite{GWSL_TPAMI}), but also flexibly adapts to various Weak supervisions. \romannumeral2) In addition, to better capture the correlation of multiple relevant labels, the network structure also contains transition loops that link different categories, which are formulated as label transition matrices. And based on the \textit{Generalized Belief Propagation} (GBP) algorithm, we derived the probability calculation algorithm for this Bayesian network.

\begin{figure}[!t]
    \centering
    \includegraphics[width=\linewidth]{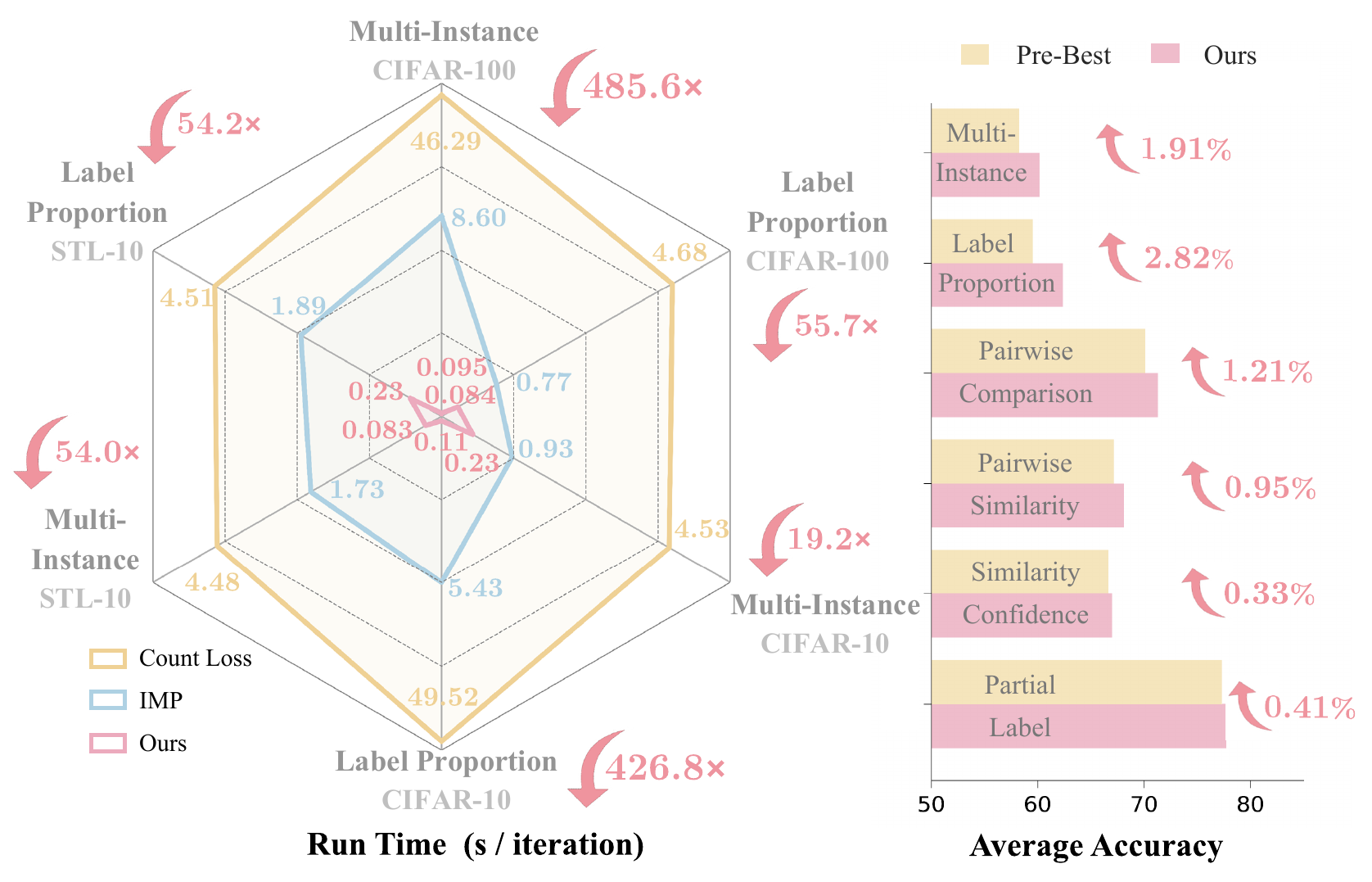}
    \caption{Comparison of run time and accuracy with recent SOTA methods on general weakly supervised learning as the batch size is 4 and the average number of instances in each bag is 20. The runtime advantage becomes more pronounced as the batch size, number of classes, or instances increases.}
    \label{fig: run time and accuracy}
\end{figure}

In order to achieve optimal time efficiency, we have explored and proposed a faster computing strategy. \romannumeral3) Firstly, after observing the sparsity of label transition probabilities, we introduce the low-rank assumption of transition matrices, thereby reducing the time complexity from quadratic to linear. Secondly, to avoid per-item (per-sample and per-category) calculation and manual hard-coding, we designed a plug-and-play and end-to-end state evolution module to learn batch-scale transition matrices, which can promote batch-wise and multi-category processing. These two improvements greatly shorten the training time, as shown in figure \ref{fig: run time and accuracy}: runtime experiments reveal consistent breakthroughs, particularly in Multi-Instance and Label Proportion Learning, achieving speedups of up to 480×. Finally, theoretically, we have also demonstrated the equivalence between our method and the EM algorithm and provided an upper bound on the generalization error. Experiments across 10+ weakly supervised settings demonstrate that our method exhibits competitive performance compared to other general or customized algorithms, achieving SOTA results in most settings.


\section{Preliminary Knowledge}
\noindent
\textbf{Supervised Learning.} Let the feature space of the instances be \(\mathcal{X}\in \mathbb{R}^{d}\), and the label space be \(\mathcal{Y}=\{0, 1\}^C\), where \(C\) represents the number of classes. In supervised learning, we have access to multiple instance-label pairs \((\bm{x}, \bm{y})\), forming a labeled dataset \(D\), where \(\bm{x}\) denotes the feature vector of an instance, and \(\bm{y}\) represents its associated multi-label annotation. The goal is to learn a multi-label classifier, obtained by minimizing the expected risk:
\begin{equation}
\label{equ: risk for supervised}
R(f) = \mathbb{E}_{(\bm{x}, \bm{y})\sim P(\bm{X}, \bm{Y})}[\mathcal{L}(f(\bm{x}), \bm{y})]
\end{equation}
where \(\mathcal{L}: \mathbb{R}^d \times \mathcal{Y}\rightarrow R\) is the loss function for supervised learning, \(\bm{X}\) and \(\bm{Y}\) are random variables of feature space and label space respectively, and \(P(\bm{X}, \bm{Y})\) is the corresponding joint distribution.

\textbf{Weakly Supervised Learning.}  In practical scenarios, obtaining an imprecisely supervised dataset \(\tilde{D}\) is relatively more feasible. Specifically, since the samples in weakly supervised datasets may consist of sets of instances (i.e. a ‘bag’), we assume that the weakly supervised dataset \(\tilde{D}\) comprises pairs of bag and weak label \((\bm{x}^{[K]}, \bm{w})\). Let each bag \(\bm{x}^{[K]}\) contain \(K\) instances, that is, \(\bm{x}^{[K]}=\{\bm{x}^k\}_{k=1}^{K}\). In addition, the weak label \(\bm{w}=(w_1, w_2, ..., w_C)\) for each bag \(\bm{x}^{[K]}\) is derived by aggregating the \textbf{unobservable} instance-level multi-labels, i.e., \(w_c=\phi(y_c^{[K]})\), where \(y^{[K]}_c=\{y_c^k\}_{k=1}^{K}\) and \(y_c^k\) is the ground-truth label of instance \(x^k\) for class \(c\),  with \(\phi(\cdot)\) representing the label aggregation rule defined by the weak supervision setting. 
Please note the following notation rule: instance level samples are represented by numerical superscripts, such as \(\bm{x}^2 \text{ and } \bm{y}^2\) representing the second instance in a bag and its label; bag-level samples use square bracket numbers as superscripts, for example, \(\bm{x}^{[6]} \text{ and } \bm{y}^{[6]}\) represent a bag including 6 instances and all instance labels within it; and the index of a category is represented by subscripts such as \(w_2\) indicating the weak label of the second class.

Using exclusively the weakly-supervised dataset \(\tilde{D}\), our goal is to construct a computationally efficient and high-performance risk estimator \(\tilde{R}(f)\), which achieves comparable effectiveness to its fully-supervised counterpart \(R(f)\), ultimately enabling the acquisition of an instance-level classifier \(f\) through systematic optimization.


    
\section{Related Work}

\subsection{Weakly-Supervised Learning} \label{sec: related1}
\noindent
Since acquiring precise annotations for fully-supervised approaches proves prohibitively costly, researchers frequently resort to constructing weakly-supervised datasets. 
In this paper, past advances on weakly-supervised learning are organized into four principal paradigms: 
\begin{IEEEitemize}[\setlength{\IEEElabelindent}{1pt}\setlength{\itemindent}{1pt}\setlength{\labelsep}{3pt}]
    \item \textbf{Imperfect supervision.}
    This paradigm is based on the premise that the supervision signals are imperfect, potentially compromising the effectiveness of model optimization. It encompasses scenarios involving (1) Noisy (\emph{Noisy}), where annotations may deviate from the ground truth with noise \cite{noisy2015, noisy2020b, noisy2020ICLR, noisy_tkde2024}; (2) Crowdsourced (\emph{CrowdL}), which involves collecting multiple noisy opinions via crowdsourcing platforms \cite{crowd2010, crowd2016, crowd2017, crowd2024xia}; (3) Partial (\emph{PartialL}), where the annotations represent a candidate label set containing the ground truth \cite{partial_2024, PICO+, partial2020NIPS, partial2017, partial2022ICML}; and (4) 
    Complementary Labeling (\emph{CompL}), in which the labels are explicitly complementary to the ground truth \cite{CompL2020}.
 
    \item \textbf{Aggregation supervision.} This type encompasses scenarios where each bag sample comprises a collection of instances, comprising (1) Multi-Instance Learning (\emph{MultiIns, MIL}), where the label indicates the presence of positive instances within one bag \cite{mil1997, 
    mil2009, mil2021, mil2021_2, multiins_2025}, and (2) Learning from Label Proportion (\emph{LProp}), which utilizes aggregated statistics such as the count or proportion of positive instances in one bag as weak labels \cite{lprop2009, lprop2020, lprop2022}.
    
    \item \textbf{Pairwise supervision.} 
    This supervision framework specifically handles cases where each bag contains exactly two instances, covering two key approaches: hard-label pairwise supervision and soft-label pairwise supervision. In hard-label settings, annotations indicate either (1) whether the first instance is more likely to be positive than the second (pairwise comparison, \emph{PairComp}) \cite{paircomp2010, paircomp2021} or (2) if the two instances are semantically similar (pairwise similarity, \emph{PairSim}) \cite{pairsim2018, GWSL_NIPS_2020, pairsim_2023}. For soft-label settings, supervision provides continuous measures such as the similarity score between instances (similarity confidence, \emph{SimConf}) \cite{simconf2021} or the difference in their confidence scores (confidence difference, \emph{ConfDiff}) \cite{confdiff2023}.
    
    \item \textbf{Incomplete supervision.} 
    This paradigm refers to learning scenarios where only a subset of the dataset is annotated, while the remaining instances lack any form of labeling. This paradigm encompasses several subcategories: positive-unlabeled learning (\emph{PosUnl}) where labeled data exclusively belong to the positive class \cite{posunl2008, posunl2015, posunl2017, posunl2022, posunl_NIPS2014}, semi-supervised learning (\emph{SemiSup}) which leverages a small labeled set with both positive and negative annotations \cite{semisup_mean_teacher, semisup_survey, semisup_gnn, semisup_MixMatch}, unlabeled unlabeled learning (\emph{UnlUnl}) where two unlabled datasets are used \cite{unlunl2019}, and similarity dissimilarity unlabeled learning (\emph{SimUnl}) which utilizes pairwise similarity or dissimilarity constraints \cite{simunl2021}. Furthermore, incomplete supervision often assumes the class prior is known or can be estimated for the unlabeled datasets.
\end{IEEEitemize}

\subsection{General Weakly-Supervised Learning} 
\noindent
While existing weak supervision paradigms (introduced earlier) have inspired specialized algorithms for distinct scenarios, their task-specific nature introduces practical limitations. Consequently, recent advances have focused on establishing general methods for weakly supervised learning.
This paper organizes these methods into two key strategies:

\subsubsection{\textbf{Inductive Properties of Base Loss}}\label{sec: Inductive Properties of Base Loss}
    CEGE \cite{GWSL_TPAMI} found that the base loss can be decomposed into label-independent and label-dependent components. The latter need to sestimate the centroid \(\sum_i{y^i \bm{x}^i}\), using two \textbf{pre-synthesized} datasets.
    Count Loss \cite{GWSL_NIPS_2023} found that the base loss \(l\) can be adapted into a count loss under certain weakly-supervised settings. 
    Let \(\hat{y}_c^k\) be the \(k\)-th instance prediction for the \(c\)-th class. Then their key step is to estimate the count probability \(P(\sum_k \hat{y}_c^k=s)\) by dynamic programming (DP) algorithms with \textbf{complexity} \(\mathcal{O}(K^2)\).

However, these methods exhibit notable limitations. CEGE \cite{GWSL_TPAMI} requires pre-synthesized datasets as the pre-work. Both CEGE \cite{GWSL_TPAMI} and Count Loss~\cite{GWSL_NIPS_2023} are tightly coupled with specific base loss functions, which may limit their applicability when the loss is altered. Additionally, none of these approaches account for multi-label learning scenarios, neglecting critical multi-label correlations.

\subsubsection{\textbf{Calculation of Latent
Probabilities} \(P(\bm{Y}^k|\bm{X}, \bm{W})\)} 
Through risk rewriting \cite{risk2022sugi}, UUM \cite{GWSL_ICML_2023} derives an unbiased risk estimator (URE) for classification from aggregate observation, which could actually be extended to other settings:

    \begin{theorem}[URE for weak supervision\cite{GWSL_ICML_2023}]\label{thm: URE}
	The risk \(\tilde{R}(f)=\mathbb{E}_{(\bm{x}^{[K]}, \bm{w})\sim P(\bm{X},\bm{W})}(\tilde{\mathcal{L}}(f_b(\bm{x}^{[K]}), \bm{w});f)\) is unbiased with respect to \(R(f)\), where
    \begin{align}
    \tilde{\mathcal{L}}&(f_b(\bm{x}^{[K]}), \bm{w};f)=\frac{1}{K}\sum_{k=1}^{K}\sum_{j=1}^C \ \\ & P(\bm{Y}^k=\bm{e}_j|\bm{X}=\bm{x}^{[K]}, \bm{W}=\bm{w})l(f(\bm{x}^k), \bm{e}_j) \nonumber
    \end{align}
    \(\bm{e}_j\) is a unit vector whose \(j\)-th element is 1, and \(f_g\) is the bag-level classifier corresponding to \(f\).
\end{theorem}

\begin{remark}
Theorem \ref{thm: URE} shows that minimizing the proposed loss \(\tilde{\mathcal{L}}\)
  on a weakly-supervised dataset \(\tilde{D}\)
  leads to a classifier consistent with those trained on fully supervised data \(D\). Additionally, URE allows for a more flexible choice of base loss \(l\), differing from the approach in Section \ref{sec: Inductive Properties of Base Loss}.
\end{remark}	

However, since the instance-level label is unobservable, \textbf{the latent probability} \(P(\bm{Y}^k|\bm{X}, \bm{W})=\frac{P(\bm{Y}^k, \bm{W}|\bm{X})}{P(\bm{W}|\bm{X})}\) is difficult to compute, thus \textbf{remaining an open problem}. We now discuss different methods to calculate it.
\begin{IEEEitemize}[\setlength{\IEEElabelindent}{1pt}\setlength{\itemindent}{1pt}\setlength{\labelsep}{3pt}]
    \item 
    GBM \cite{GWSL_NIPS_2020} and UUM \cite{GWSL_ICML_2023} adopted a \textbf{direct computation} approach, where (random vars. omitted for simplicity):
    \begin{equation} 
        \label{equ: UUM direct computation}
        P(\bm{w} \mid \bm{x}^{[K]}) = \sum_{\bm{y}^{[K]} \in \sigma(\bm{w})} \prod_{k=1}^K P(\bm{y}^k \mid \bm{x}^{[K]})
    \end{equation}
    and \(\sigma(\bm{w})\) denotes the set of all instance-level labels that generate the weak label \(\bm{w}\). Crucially, direct computation requires enumerating all possible instance-level label configurations, with \textbf{worst-case time complexity \(\mathcal{O}(C^K)\)}.

    \item 
    For different weak supervision scenarios, GLWS \cite{GWSL_ICML_2024} \textbf{predefines} the process of \(\bm{y}^{[K]}\) generating \(\bm{w}\) as \textbf{different nondeterministic finite automatons (NFA)}, expands it into a grid graph, and then uses the forward-backward algorithm to calculate \(P(\bm{Y}^k, \bm{W}|\bm{X})\). The algorithm has a \textbf{time complexity} of \(\mathcal{O}(K|W|^2)\), where \(|W|\) represents the number of values of the weak supervision label.
\end{IEEEitemize}
In summary, UUM \cite{GWSL_ICML_2023} suffers from high complexity, while GLWS \cite{GWSL_ICML_2024} customizes NFAs for different weak labels. This requires case-by-case programming with even per-sample and per-category computations (e.g., in \emph{MultiIns} and \emph{LProp}), instead of efficient batch processing (simultaneously computing across all categories for entire batches). Moreover, both approaches fail to model multi-label dependencies.

\section{Methodology}
\noindent
As noted, the mentioned method for latent probability calculation still faces three key issues in Table \ref{tab: related work}, addressed in the subsections:
\romannumeral1) To avoid complex manual pre-work, we introduce a shared unified Bayes Network structure for various weak supervisions.
\romannumeral2) To model the overlooked multi-label dependencies, we formulate the label transition probabilities and derive the GBP-based algorithm for probability calculation.
\romannumeral3) To reduce the time complexity compared to prior studies, we design the accelerated algorithm.
\romannumeral4) To enable batch processing, we propose a new module with an auxiliary loss, eliminating per-sample and per-category probability calculations \cite{GWSL_ICML_2024}. 
The last two points significantly achieve the speedup.

\subsection{Bayesian Network with Loop}
\noindent
As discussed earlier, the main challenge of developing general weakly supervised algorithms is how to estimate the latent probability \(P(\bm{Y}| \bm{X}^{[K]}, \bm{W})\). Actually, previous methods attempt to break down the process of inferring the possible instance-level labels \(\bm{Y}\) from weak labels \(\bm{W}\) by modeling it using different graph structures: 
\begin{IEEEitemize}[\setlength{\IEEElabelindent}{1pt}\setlength{\itemindent}{1pt}\setlength{\labelsep}{3pt}]
    \item GLWS \cite{GWSL_ICML_2024} pre-design weak supervisions as varied NFAs.
    \item UUM \cite{GWSL_ICML_2023} directly enumerates possible instance labels using brute-force algorithms like Depth-First Search (DFS).
\end{IEEEitemize}
For example, figure \ref{fig: DFS} shows the DFS process for a single category in \emph{MultiIns} when the number of instances in one bag is \(3\), where \(Z^k\) indicates whether there is a positive sample in the first \(k\) instances, and \(Y^k\) is the label of the \(k\)-th instance. According to the values of \(Z^{k-1}\) and \(Y^{k-1}\), traversing the next instance label \(Y^k\) can search for all the cases of \(Z^k\) and \(Y^k\) in the next layer. All the instance label solutions can be obtained by simply recording the search path in the DFS structure.


\subsubsection{\textbf{Structure of Graphical Model}} \label{subsubsec: Structure of Graphical Model}
\begin{figure}[!t]
    \centering
    \includegraphics[width=0.88\linewidth]{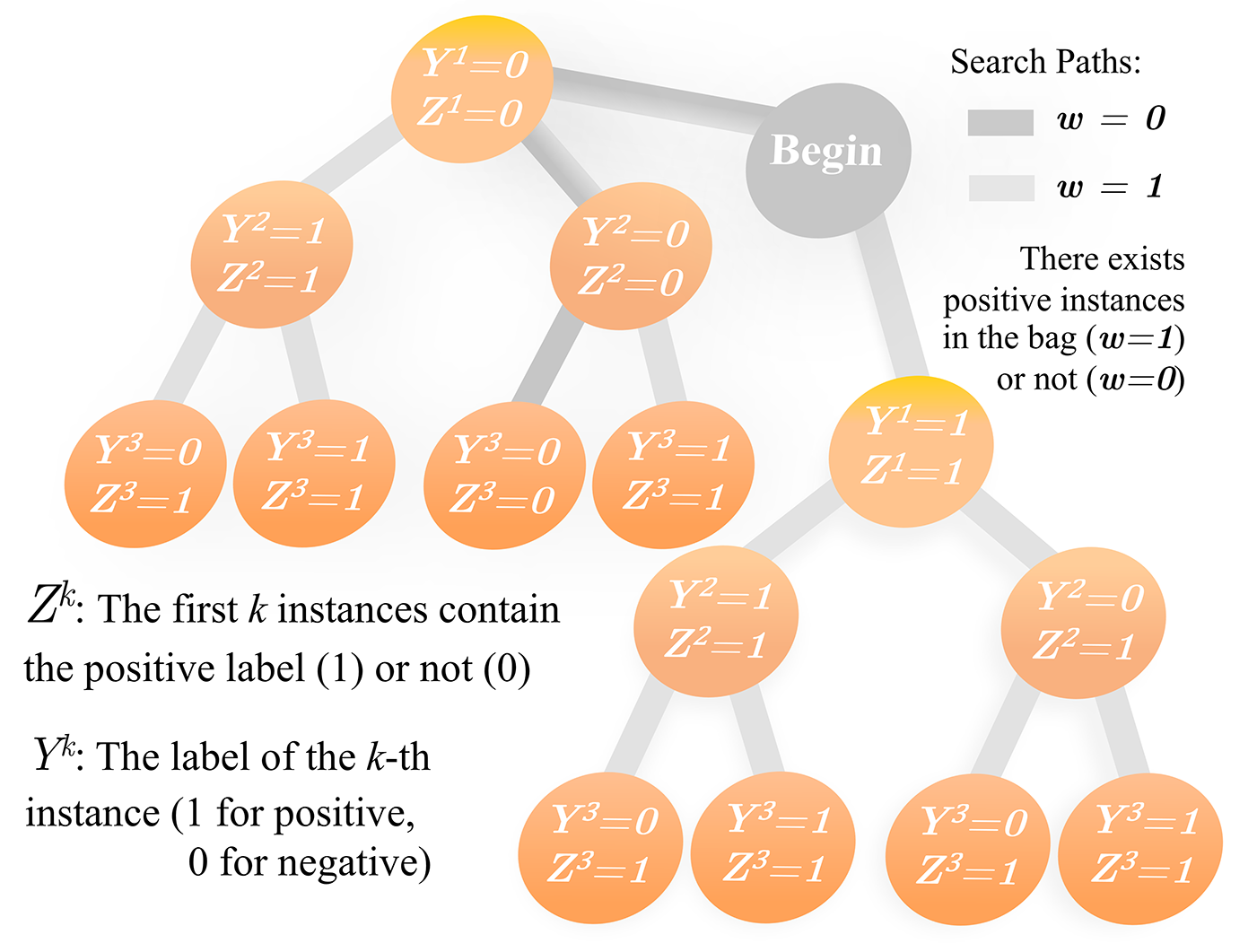}
    \caption{DFS tree for Multi-Instance Learning when the number of instances in one bag is 3 for only one class. Note that nodes of the same color belong to the same layer.}
    \label{fig: DFS}
\end{figure}

However, both methods remain neither a unified nor an efficient strategy due to two key limitations: \romannumeral1) The computational overhead of DFS is excessively high, as its time complexity grows exponentially; \romannumeral2) As illustrated in figure \ref{fig: DFS}, the search paths of graphs could vary when weak labels differ. In practice, mini-batch data typically contains inconsistent labels, and for multi-label learning, even the labels of different categories for the same sample can vary, forcing them to adopt sample- and category-specific computation while preventing batch processing. 


Inspired by previous work \cite{GBP1, GBP2, GBP3, GWSL_ICML_2024}, graphical model 
may resolve both limitations mentioned above: \romannumeral1) graphical models achieve superior time complexity compared to DFS; \romannumeral2) graph structures under different weak labels just differ in the reachability of nodes, while unreachable nodes could be exactly encoded as zero-probability states using graphical model algorithms. By applying this rule, all weak labels could share identical graph structures, differing only in node probability distributions. Thus, we transform the mining process into a probabilistic graph model. 
In figure~\ref{fig: DFS}, by treating the parameters \(Z^k\) and \(Y^k\) as random variables, nodes at identical tree layers (with identical superscripts) can be merged into random vector nodes \((Y^k, Z^k)\), which converts the DFS tree (e.g., the orange one in figure \ref{fig: DFS}) into a chain-structured probabilistic graphical model for one class (e.g., the orange chain in figure \ref{fig: bayes loop}). 
\begin{table}[!t]
\centering
\caption{Meanings of Node Variables in DFS Tree or Bayesian Network for Different Weak Supervision}
\label{tab: node meaning}
\begin{tabular}{@{}>{\centering\arraybackslash}m{3cm} 
                >{\centering\arraybackslash}m{5cm}@{}}
\toprule
\multicolumn{1}{c}{\textbf{Weak Supervision}}                                                  
    & \multicolumn{1}{c}{\textbf{Meanings of \(Z^k\)}}  \\
\midrule
\emph{LProp}, \emph{PosUnl}, \emph{UnlUnl} 
    & number of positive samples for the first k instances in the bag~~       \\
\emph{PairComp}                                                               
    & more positive than the first instance or not        \\
\emph{PairSim}                                                                
    & similar to the first instance or not        \\
...                                                                                            
    & ...                                  \\
\emph{PartialL}                                                               
    & the k-th category is considered as a candidate label or not   \\
\bottomrule
\end{tabular}
\end{table}

However, figure~\ref{fig: DFS} illustrates the DFS tree for only one category, while our focus lies in multi-label learning (MLL) with inherent correlations among labels. These correlation presents a key challenge in MLL, like identifying dependencies between symptoms (fever/headache) or mislabeling of visual features (zebra/white tiger fur) \cite{crowd2025}. And graphical models handle this by encoding correlations through cyclic structures, forming a \textbf{loopy Bayesian network} in figure \ref{fig: bayes loop} where interconnected loops establish label-chain correlation – generalizing the single-chain NFAs \cite{GWSL_ICML_2024} as a special case.

To realize probability inference in such structures, we employ the Generalized Belief Propagation (GBP) algorithm \cite{GBP1, GBP2, GBP3} in the following sections. Notably, the method in GLWS \cite{GWSL_ICML_2024} fails to handle loopy graph topologies.

\subsubsection{\textbf{Advantages and Novelties}} 
\begin{figure*}[!t]
    \centering
    \includegraphics[width=0.9\linewidth]{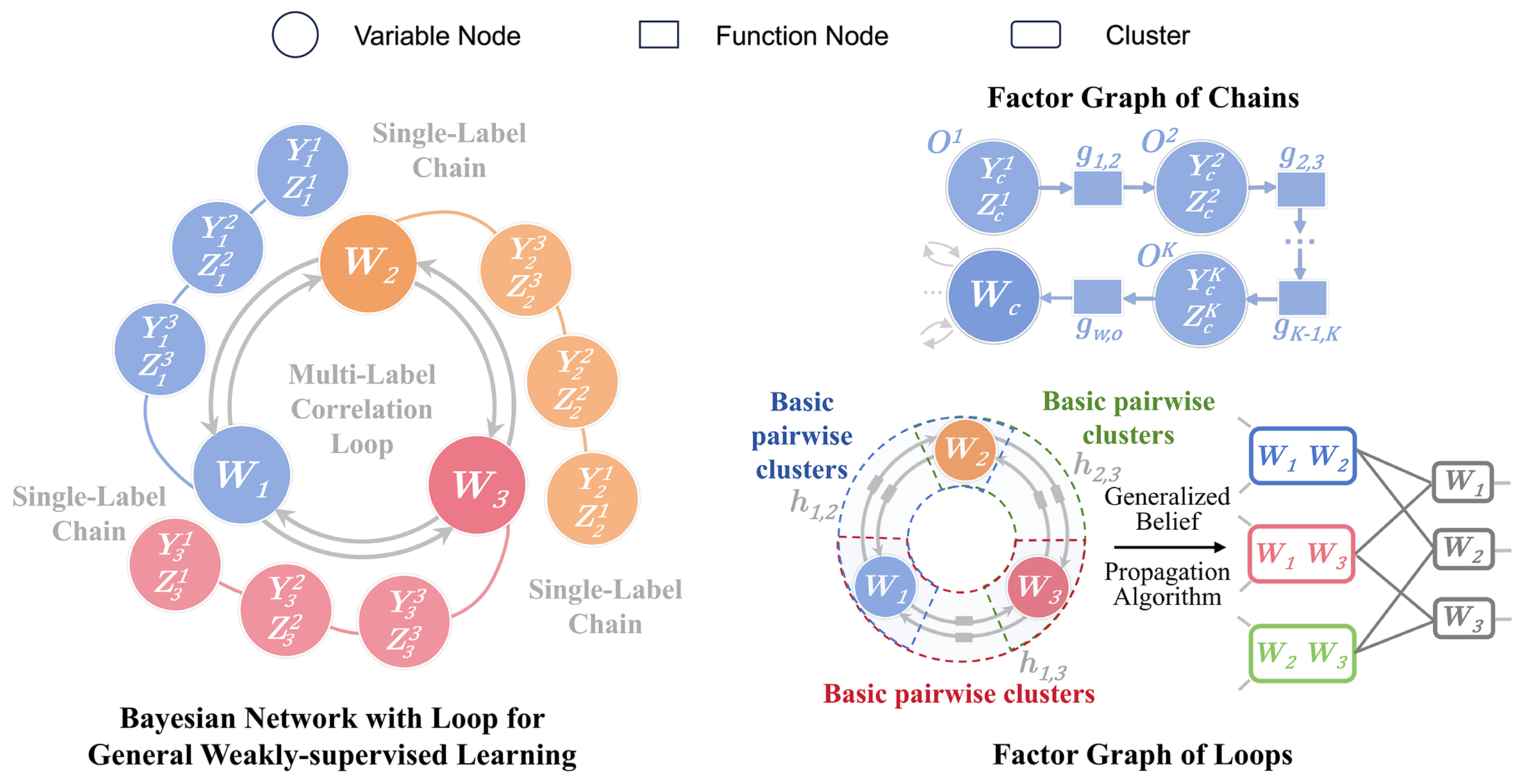}
    \caption{A loopy Bayesian network model for general weak supervision. Different colors denote distinct label types. (a) Left panel: Visualization of a probabilistic graph with 3 instances and 3 labels, where all feasible instance configurations are mined through chain-structured dependencies, while loops explicitly characterize multi-label correlations. Right panel: Factor graph representation corresponding to the chain (b) and loop (c) components of the left network, employing generalized belief propagation (GBP) for efficient probabilistic inference.}
    \label{fig: bayes loop}
\end{figure*}

\paragraph{Multi-Label Correlation} 
The loopy structure of the Bayesian network (figure \ref{fig: bayes loop}) encodes multi-label dependencies by coupling inter-class generative processes. Unlike previous works that assume class independence, our method could better address this challenge in MLL.
\paragraph{No Pre-Work} 
Our Bayesian framework, derived from DFS trees, also processes the brute-force nature of DFS, which can be readily extended to various weakly supervised settings beyond Multi-Instance Multi-Label. As shown in Table \ref{tab: node meaning}, only the interpretation of \(Z^k\) varies across settings: it may denote the number of positive instances among the first \(k\) samples (in \emph{LProp}, \emph{PosUnl}, and \emph{UnlUnl}), or indicate whether the current instance is more likely to be positive than the first one (in \emph{PairComp}), among others. Therefore, our unified Bayesian structure offers flexible adaptation to diverse weakly supervised scenarios, eliminating the need for customized NFAs designed for each weak supervision \cite{GWSL_ICML_2024} or pre-synthesized datasets \cite{GWSL_TPAMI}.

\subsection{Probability Calculation}
\noindent
Following the definition of the general structure for weak supervision, the next step involves computing latent probability within the Bayesian network via the GBP algorithm. The GBP algorithm \cite{GBP1, GBP2, GBP3} necessitates transforming the original graph into a factor graph \cite{factorgraph} by adding function nodes, as illustrated in the right part of figure \ref{fig: bayes loop}. 

\subsubsection{\textbf{Calculation Process}} 
\textbf{Notations}. For nodes on chains, the random vector nodes \(\bm{O}^i=(Y^i_{\cdot}, Z^i_{\cdot})\) is abbreviated, with their values \((y^i, z^i)\) denoted as \(\bm{o}^i\). The set of all possible values of \(\bm{o}^i\) is set to \(S_o\), where typically \(|S_o|=|W|\cdot |Y|\). The function node between \(O^{i-1}\) and \(O^i\) is represented by \(g_{i-1, i}\). For loops, consistent with previous content, the weak label node for the \(c\)-the class is denoted as \(W_c\). And the function node between \(W_i\) and \(W_j\) is represented by \(h_{i,j}\). Finally, the message passed from node \(\square\) to \(\triangle\) is defined as \(\mu_{\square\rightarrow \triangle}\).
     
\textbf{Chains}. The function nodes \(g_{i-1,i}\) on chains could be understood as transition probabilities \(P(\bm{O}^i=\bm{o}^i| \bm{X}^{[K]}=\bm{x}^{[K]}, \bm{O}^{i-1}=\bm{o}^{i-1})\). For the convenience of explanation, we formulate them into matrix form
\(
    \bm{T}_{k,k+1}(\bm{x}^{[K]})=(P(\bm{O}^{k+1}=\bm{o}^{k+1}|\bm{O}^{k}=\bm{o}^k, \bm{X}^{[K]}=\bm{x}^{[K]}))_{\bm{o}^{k+1},\bm{o}^{k}\in S_o}\in \mathbb{R}^{|S_o|\times |S_o|}
\). Then the structure of the transition matrices
\(\bm{T}_{k, k+1}(\bm{x}^{[K]})\) is ideally similar to the adjacency matrix of the DFS tree. For example, in \emph{MultiIns} like figure \ref{fig: DFS}, the DFS adjacency matrix becomes (the state (1,0) does not exist, making the third column arbitrary):
\begin{equation} \label{equ: adjacency matrix}
\centering  
\vcenter{\hbox{
\begin{tikzpicture}[
  mymatrix/.style={
    matrix of math nodes,
    nodes in empty cells,
    left delimiter={(},
    right delimiter={)},
    nodes={minimum width=5ex, minimum height=3ex, anchor=center},
    column sep=0.5ex,
    row sep=0.5ex
  }
]
\matrix[mymatrix] (mat) {
1 & 0 & 1 & 0 \\
0 & 1 & 0 & 1 \\
0 & 0 & 0 & 0 \\
1 & 1 & 1 & 1 \\
};

\draw[dashed, line width=0.4pt] 
  ($(mat-1-2.north east)!0.5!(mat-1-3.north west)$) -- 
  ($(mat-4-2.south east)!0.5!(mat-4-3.south west)$);

\node[above=2.0ex of mat.north] (toplabel) {\((Y^k, Z^k)\)};
\node[above left=-0.5ex and 1.5ex of mat.north west, anchor=south east] (leftlabel) {\((Y^{k+1}, Z^{k+1})\)};

\node[left=of mat-1-1.west, anchor=east] (row1) {\((0,0)\)};
\node[left=of mat-2-1.west, anchor=east] (row2) {\((0,1)\)};
\node[left=of mat-3-1.west, anchor=east] (row3) {\((1,0)\)};
\node[left=of mat-4-1.west, anchor=east] (row4) {\((1,1)\)};
\node[above=0.05ex of mat-1-1.north] (col1) {\((0,0)\)};
\node[above=0.05ex of mat-1-2.north] (col2) {\((0,1)\)};
\node[above=0.05ex of mat-1-3.north] (col3) {\((1,0)\)};
\node[above=0.05ex of mat-1-4.north] (col4) {\((1,1)\)};
\end{tikzpicture}
}}
\end{equation}
Additional settings are in the appendices. In terms of differences, adjacency matrix entries with value 1 exhibit non-zero probabilities in \(\bm{T}_{k,k+1}(\bm{x}^{[K]})\), while others are zero. These values of function nodes encode reachability (whether transitions exist) and likelihood weights (transition probabilities) along paths in Bayesian networks of figure \ref{fig: bayes loop}.

\textbf{Loops}.
The function nodes on loops are exactly set to the symmetric class transition probability \(h_{i, j}(w_i, w_j)=P(W_i=w_i|W_j=w_j,\bm{X}^{[K]}=\bm{x}^{[K]})=P(W_j=w_j|W_i=w_i,\bm{X}^{[K]}=\bm{x}^{[K]})\), widely studied in weakly supervised learning \cite{noisy2015, TransMatix2017, TransMarix2022, TransMatrix2023}. It usually represents the correlation between labels in a probabilistic form. Specifically, for sample \(\bm{x}^{[K]}\), the \(i\)-th category will be labeled as the \(j\)-th category with a probability of \(P(W_j=w_j|W_i=w_i,\bm{X}^{[K]}=\bm{x}^{[K]})\), which has been ignored in previous studies \cite{GWSL_ICML_2023, GWSL_ICML_2024, GWSL_NIPS_2023, GWSL_TPAMI}.

We now derive the probability calculation conclusion based on the GBP algorithm \cite{GBP1, GBP2, GBP3}. Initialize the probability of the first node to \(\bm{O}^1=\bm{o}^1\) to \(P(\bm{O}^1=\bm{o}^1|\bm{X}^{[K]}=\bm{x}^{[K]})\), and set \(g_{w, o}\) to \(\frac{P(\bm{O}^K=\bm{o}^{K}|W=w,\bm{X}^{[K]}=\bm{x}^{[K]})}{P(\bm{O}^K=\bm{o}^K|\bm{X}^{[K]}=\bm{x}^{[K]})}\). Then we have:


\begin{theorem}[Belief of the nodes in figure \ref{fig: bayes loop}]\label{thm: chain}
For each node \(\bm{O}^k\) in the chain, its belief becomes
\begin{equation}
    \begin{aligned}
    b_{\bm{O}^k}(\bm{o}^k)&=Z^{-1}\prod_{g\in n(\bm{O}^k)} \mu_{g\rightarrow \bm{O}^k}(\bm{o}^k)  \\
    & =P(\bm{O}^k=\bm{o}^k| W=w, \bm{X}^{[K]}=\bm{x}^{[K]})  
    \end{aligned}
    \label{equ: belief chain}
\end{equation}
where \(n(\bm{O}^k)\) represents the neighbor node of \(\bm{O}^k\), \(\mu_{g\rightarrow \bm{O}^k}\) is the message passing from \(g\) to \(\bm{O}^{k}\), \(b_{\bm{O}^k}(\cdot)\) is the belief of node \(\bm{O}^k\) and \(Z\) is the normalization factor.

For each node in loops, its belief \(b_{W_i}(w_i)\) becomes:
    \begin{equation}
        \begin{aligned}
            b_{W_i}(w_i) &=\prod_{h \in n(W_i)} \mu_{h\rightarrow W_i}(w_i) \\
            & = P(W_i=w_i | \bm{X}^{[K]}=\bm{x}^{[K]})  
        \end{aligned}
        \label{equ: belief loop}
    \end{equation}
    where, for \(k\in[C] / \{i\}\), we have \\ \(\mu_{h_{k, i}\rightarrow W_i}( w_i)=\sum_{w_k}P(W_i=w_i|W_k=w_k, \bm{X}^{[K]}=\bm{x}^{[K]})\mu_{g_{w_k, o_k^K}\rightarrow W_k}(w_k)\)
\end{theorem}
\begin{remark}
    Theorem \ref{thm: chain} indicates that the latent probability of the random vector can be directly obtained from the node's belief. Therefore, the \textbf{latent probability} of \(Y^k\) can be computed by summing over all possible states of \(Z^k\).
    i.e.,
\(P(Y^k=e|\bm{X}^{[K]}=\bm{x}^{[K]}, W=w)=\sum_{z^k:\bm{o}^k=(e, z^k)}b_{\bm{O}^k}(\bm{o}^k)\).

    The calculation involving \(P(W_i|W_j, \bm{X}^{[K]})\) within loop structures effectively models dependencies among categories. This capability supports applications in noisy label and multi-label learning, not adequately handled in prior approaches.
\end{remark}


\subsubsection{\textbf{Advantages and Novelties}}

\paragraph{Multi-Label Correlation} We capture state transitions and multi-label correlations through transition probabilities, i.e., \(P(\bm{O}^i=\bm{o}^i | \bm{O}^{i-1}=\bm{o}^{i-1}, \bm{X}^{[K]}=\bm{x}^{[K]})\) and \(P(W_j=w_j|W_i=w_i,\bm{X}^{[K]}=\bm{x}^{[K]})\). This is a key challenge in MLL but has been ignored by previous work.

\subsection{Single-Step Acceleration}
\noindent
The transition matrices \(\bm{T}_{k, k+1}\) structurally resemble the adjacency matrix like Equation (\ref{equ: adjacency matrix}), which is sparse and not full rank. In this section, we leverage these properties to further optimize the calculation process.

To present clearly, we vectorize the message \(\mu_{g\rightarrow \bm{O}^k}\) in Theorem \ref{thm: chain}
by defining \(\bm{\mu}_{k}^{(1)}=(\mu_{g_{k-1, k}\rightarrow \bm{O}^{k}}(\bm{o}^{k}))_{\bm{o}_k\in S_o}\) and \(\bm{\mu}_{k}^{(2)}=(\mu_{g_{k, k+1}\rightarrow \bm{O}^{k}}(\bm{o}^{k}))_{\bm{o}^{k} \in S_o}\). This allows reformulating the message passing rule as matrix operations: 
\begin{equation}
    \begin{aligned}
    \bm{\mu}_{k+1}^{(1)}&=\bm{T}_{k, k+1}(\bm{x}^{[K]})\bm{\mu}_{k}^{(1)} \\ \bm{\mu}_{k}^{(2)}&=\bm{T}^T_{k, k+1}(\bm{x}^{[K]})\bm{\mu}_{k+1}^{(2)} 
    \end{aligned}
    \label{equ: message0}
\end{equation}

\subsubsection{\textbf{Low-Rank Assumption}}
As shown in the adjacency matrix (Equation (\ref{equ: adjacency matrix})), the columns on the left and right side of the dotted line are identical. This column repetition indicates that the matrix is rank-deficient (not full rank, and even low-rank). We provide additional examples illustrating low-rank matrices for other weakly supervised scenarios in the supplementary material. Given the structural similarity between the adjacency matrix and the transition matrix, we hypothesize that the transition matrix is also of low rank.

\begin{assumption}[Low rank properties]\label{asp: low rank}
The transition matrices \(\bm{T}_{k, k+1}(\bm{x}^{[K]})\) in Equation (\ref{equ: message0}) for Theorem \ref{thm: chain} is of low rank, so that \(\exists\ \bm{U}_{k, k+1}(\bm{x}^{[K]})\in \mathbb{R}^{|S_o|\times r}\), and \(\bm{V}_{k,k+1}(\bm{x}^{[K]})\in \mathbb{R}^{|S_o|\times r}\), where \(r \ll |S_o|, s.t. \)
\begin{align}
    \bm{T}_{k, k+1}(\bm{x}^{[K]})= \bm{U}_{k,k+1}(\bm{x}^{[K]})\cdot \bm{V}_{k, k+1}^T(\bm{x}^{[K]}) \nonumber
\end{align}
\end{assumption}
In this case, the message passing rule in Equation (\ref{equ: message0}) can be re-expressed as a lower time complexity form:
\begin{equation}
    \begin{aligned}
    \bm{\mu}_{k+1}^{(1)}&=\bm{U}_{k, k+1}(\bm{x}^{[K]})\cdot             
        (\bm{V}_{k, k+1}^T(\bm{x}^{[K]})\bm{\mu}_{k}^{(1)})  \\
    \bm{\mu}_{k}^{(2)}&=\bm{V}_{k, k+1}(\bm{x}^{[K]}) \cdot             
        (\bm{U}_{k, k+1}^T(\bm{x}^{[K]})\bm{\mu}_{k+1}^{(2)}) 
    \end{aligned}
    \label{equ: message1}
\end{equation}
In the implementation details, we set the matrix \(V\) to be the first \(r\) rows of the two identity matrices, i.e.
\begin{equation}
	\bm{V}_{k, k+1}^T(\bm{x}^{[K]}) \equiv 
	\begin{pmatrix}
		I_{|W|} & I_{|W|}
	\end{pmatrix}_{:r, :}
	\label{equ: matrix V}
\end{equation}
where \(I\) is the identity matrix, and \(:r, :\) denotes the first r rows. 
Equation \eqref{equ: adjacency matrix} shows that the identity matrices could replicate the left partition to the right to approximate the adjacency structure.

Algorithm \ref{alg: calculation} illustrates the GBP calculation process. Firstly, following Theorem \ref{thm: chain}, we pass messages item by item along the chain. Subsequently, Theorem \ref{thm: chain} is applied to compute the message of weakly labeled nodes on loops, involving inter-class transitions. After messages are propagated back from the loop, all information received by each node is aggregated to derive the latent probability.
 Note that the message passing rule \(\mu_{\cdot \rightarrow W_i}\) in the loop indicates that we do not treat different labels independently, but model their relationship through the label transition probability. 
 In implementation details, if no multi-label dependencies exist, just disable the gradient of probability \(P(W_i|W_k, \bm{X}^{[K]})\).

 \begin{figure}[!t]
    \centering
        \begin{minipage}{0.95\linewidth}
        \begin{algorithm}[H]
            \caption{GBP Calculation Process.}
            \label{alg: calculation}
            \begin{algorithmic}[1]
        \REQUIRE  The prediction \(\bm{p}\) from backbone,  bag-level labels \(\bm{w}\), matrices \(\bm{T}_{k, k+1}\) or \(\bm{U}_{k, k+1}\), and inter-class transition matrix \(\bm{T}_{class}\) for loopy categories dependencies, composing \(P(W_i=w_i| W_j=w_j, \bm{X}^{[K]}=\bm{x}^{[K]})\).
        
        \STATE Initial the message \(\bm{\mu}^{(1)}_1\) for the first node.
        \STATE According to Equation (\ref{equ: message0}) or its low-rank form (\ref{equ: message1}), compute the message \(\bm{\mu}^{(1)}_k, \text{for } k=1,\ ...,K-1\).
        
        \STATE Calculation the message sent to weak label node \(\bm{W}\) using Theorem \ref{thm: chain} with matrix \(\bm{T}_{class}\). In particular, 
        \begin{equation}
            \bm{\mu}_{\cdot \rightarrow \bm{W}} = g_{w, o}(\bm{\mu}^{(1)}_K{}) \cdot \bm{T}_{class} \nonumber
        \end{equation}
        \algorithmiccomment{The matrix multiplication involving \(\bm{T}_{class}\) models the multi-label association relationship.}
        
        \STATE Initial the message \(\bm{\mu}^{(2)}_K\) for the last node on chains.
        \STATE According to Equation (\ref{equ: message0}) or (\ref{equ: message1}), compute the message \(\bm{\mu}^{(2)}_k, \text{for } k=K-1,K-2, ..., 1\).
        
        \STATE Aggregate the node messages to form beliefs \(b_{\cdot}(\cdot)\) and calculate the latent probability \(P(\bm{Y}| \bm{X}^{[K]}, \bm{W})\).
        \end{algorithmic}
    \end{algorithm}
    \end{minipage}
    \end{figure}

\subsubsection{\textbf{Advantages and Novelties}}
\paragraph{Time Complexity}
As summarized in Table \ref{tab: related work}, Count Loss \cite{GWSL_NIPS_2023} formulates probability computation as a two-dimensional DP problem, resulting in a time complexity quadratic with respect to the number of instances in the bag, i.e., \(\mathcal{O}(K^2)\); UUM employs an exhaustive enumeration strategy, which leads to a time complexity of \(\mathcal{O}(C^K)\). GLWS \cite{GWSL_ICML_2024} models different weak supervisions using NFAs and applies the forward-backward algorithm for probability computation, with a time complexity that is linear in the number of instances and quadratic in the number of node states, i.e., \(\mathcal{O}(K|W|^2)\). For our algorithm, since \(r \ll |S_o|\),
the time complexity of each step in Equation (\ref{equ: message1}) is \(O(2|S_o|\times r)=O(2|S_o|)\), and a total of \(K\) steps are required (K instances, K layers of DFS trees). Therefore, the overall time complexity is \(O(K|S_o|)\). And the node \(\bm{O}^i=(Z^i, Y^i), Y^i \in\{0, 1\}\) indicates the number of values of \(Y^i\) is limited, so the time complexity is reduced to \(O(K|W|)\).

\subsection{Batch Learning}
\noindent
As mentioned in Section \ref{subsubsec: Structure of Graphical Model}, the weak labels of bags in one batch remain inconsistent, so search paths vary significantly across these bags. Some methods \cite{GWSL_ICML_2024, GWSL_NIPS_2023} manually define and hard-code probabilistic transition rules for each sample and category, which is highly time-consuming. To address this, as suggested by Equation (\ref{equ: message1}) in this paper, if the matrix \(\bm{U}\) is batched, it would no longer be necessary to iterate through all samples and categories within the batch. Using a neural network (NN) to learn such a batched matrix is a feasible approach for two main reasons:
	\begin{IEEEitemize}[\setlength{\IEEElabelindent}{1pt}\setlength{\itemindent}{1pt}\setlength{\labelsep}{3pt}]
		\item By constraining the output range of NNs, zero values can represent non-existent states, while non-zero probabilities can indicate the possibility of state transitions.
		\item The matrix learned by NNs can be easily designed in a batched form, enabling simultaneous computation across all samples and categories within the batch.
	\end{IEEEitemize}

While a standard MLP could partially accomplish this task, it still faces two major limitations: first, the probability values in the transition matrix \(\bm{U}\) are sparse, whereas MLP outputs tend to be dense and unbounded; second, the transition from previous-layer nodes to the next layer resembles a process of querying historical state features using instance features to determine transition feasibility—a mechanism analogous to attention, which is difficult for standard MLPs to capture. To overcome these issues, we introduce in this section a novel plug-and-play module to facilitate the learning of the batched transition matrix \(\bm{U}\).

\subsubsection{\textbf{State Evolution Module}}
\begin{figure}[!t]
    \centering
    \includegraphics[width=0.86\linewidth]{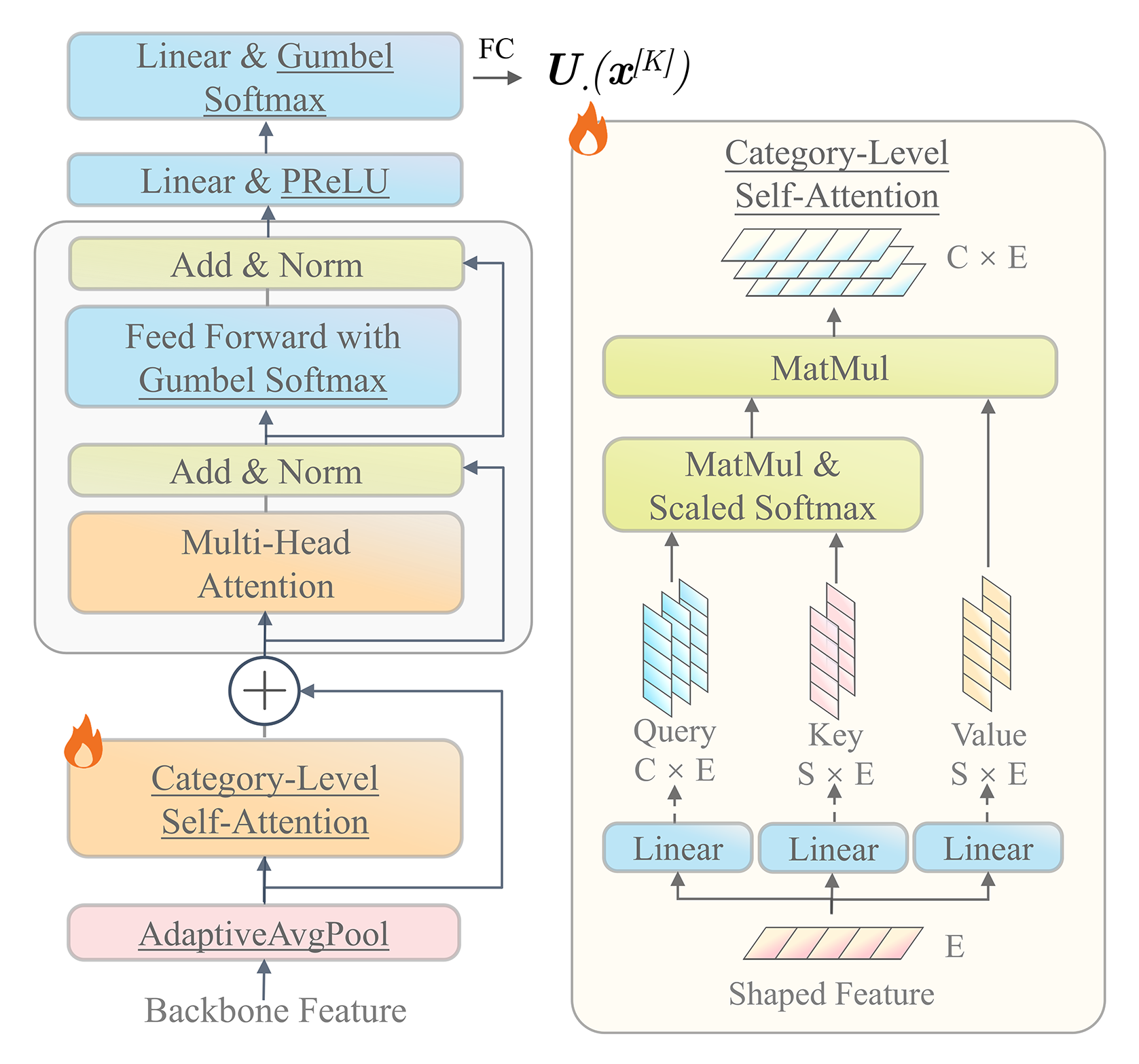}
    \caption{The end-to-end state evolution module to learn transition matrices \(\bm{U}(\bm{x}^{[K]})\). The core innovation points are marked with underlines.}
    \label{fig: matrix learner}
\end{figure}

Figure \ref{fig: matrix learner} shows our proposed state evolution module, which can be integrated into other backbones and learn batched transition matrices \(\bm{U}\) end-to-end.
First, features \(\bm{\mathscr{F}}' \in \mathbb{R}^{B\times L \times E'}\) from the backbone are compressed via adaptive average pooling into \(\bm{\mathscr{F}} \in \mathbb{R}^{B\times L \times E}\) where \(B\) is the batch size, \(L\) is the maximum length of bags in a mini-batch and \(E'\) (\(E\)) is the (reshaped) feature size. It then passes through a category-level self-attention module to capture dependencies between categories and instances:

\begin{equation}
    \begin{aligned}
        \bm{\square} &= \text{Reshape}(\text{Linear}_{\square}(\bm{\mathscr{F}}))\  \text{where} \ \square \in \{\bm{\mathscr{Q}}, \bm{\mathscr{K}}, \bm{\mathscr{V}}\}\}\\
        \bm{\mathscr{A}} &= \text{Softmax}(\frac{\text{MatMul}(\bm{\mathscr{Q}}, \bm{\mathscr{K}}^T)}{\sqrt{E}})
    \end{aligned}
\end{equation}
where \(\bm{\mathscr{Q}} \in \mathbb{R}^{B\times L \times C\times E}\), \(\bm{\mathscr{K}}, \bm{\mathscr{V}} \in \mathbb{R}^{B\times L \times S \times E}, \bm{\mathscr{A}} \in \mathbb{R}^{B\times L \times C\times S}\), and \(S\) is the number of keys. 
In this step, the attention weights are computed as the normalized inner-product similarity between each category feature and \(S\) instance-level feature vectors. These weights primarily indicate whether the instance features support the state transition of different categorical nodes in the graphical model.

A residual aggregation produces transition-aware features between the original input \(\bm{\mathscr{F}}\) and attention weighted value \(\bm{\mathscr{A}} \cdot \bm{\mathscr{V}}\). These features are then processed by a transformer encoder \cite{transformer} using Gumbel-Softmax (GS) \cite{gumbel_softmax_E} with temperature \(\tau=1.0\) as the activation function:
\begin{equation}
	\text{GS}(\bm{x})=\text{Softmax}(\frac{\bm{x} - \log(-\log \bm{\epsilon})}{\tau}),\ \bm{\epsilon} \sim \mathcal{U}(0, 1)
\end{equation}
On one hand, the GS function introduces randomness, simulating the noise in data annotation and improving the generalization capabilities. On the other hand, GS functions encode the connection information through 0-1 outputs.

Finally, to approximate the discrete adjacency structure, two linear layers with PReLU and Gumbel-Softmax (\(\tau=0.5\)) activations are applied to generate a sparse matrix \(\bm{U}\). 
 In implementation details, masking \(\bm{U}\) with the adjacency matrices could further speed up convergence.

\subsubsection{\textbf{Auxiliary Loss}}
As for the training pattern, we use a one-stage strategy to jointly optimize the backbone and the proposed module. Specifically, alongside the unbiased loss \(\tilde{\mathcal{L}}\) in Theorem \ref{thm: URE}, a latent probability smoothing loss \(\mathcal{L}_{smo}\) is also introduced:
\begin{equation}
\begin{aligned}
    &\mathcal{L}_{total}(f(\bm{x}^{[K]}), \bm{w})\\ =&\tilde{\mathcal{L}}(f(\bm{x}^{[K]}), \bm{w}) + \lambda  \mathcal{L}_{smo}(f(\bm{x}^{[K]}), \bm{w}) \\
    = &\tilde{\mathcal{L}}(f(\bm{x}^{[K]}), \bm{w}) + \lambda\cdot [\frac{1}{CK}\sum_{k=1}^K \sum_{j=1}^C  \\& \ \ P(\bm{e_j}|\bm{x}^{[K]}, \bm{w})  log P(\bm{e_j}|\bm{x}^{[K]}, \bm{w})] 
\end{aligned}
\end{equation}
where \(P(\bm{e_j}|\bm{x}^{[K]}, \bm{w})\) is short for \(P(\bm{Y}^k=\bm{e_j}|\bm{X}^{[K]}=\bm{x}^{[K]}, \bm{W}=\bm{w})\), and \(\lambda\) is the balanced coefficient. 

\(\mathcal{L}_{smo}\) acts as a negative entropy regularizer, preventing long-tailed latent distributions and overfitting to noise, especially early in training, thus speeding up convergence.

When the base loss \(l\) is binary cross-entropy (BCE) and \(\lambda=1\), \(\mathcal{L}_{total}\) simplifies to the Kullback–Leibler divergence. This minimizes the discrepancy between the module’s latent probability and the backbone’s prediction, thereby reducing prediction bias due to distribution misalignment.

\subsubsection{\textbf{Advantages and Novelties}}
\paragraph{Time Complexity} 
Compared to prior methods, our approach achieves the best complexity \(\mathcal{O}(K|W|)\). Although it only reduces by one polynomial order, this yields a significant practical speedup, due to three innovations: \romannumeral1) Low-rank matrix approximations optimize message passing processes. \romannumeral2) Sparse matrix multiplication could also accelerate the computation. \romannumeral3) An end-to-end module learns the batch-wise, multi-category transition matrices \(\bm{U}\), enabling efficient cross-category batch processing without per-instance and pre-category calculation.

\subsection{Further Analysis}
\noindent
In weak-supervised learning, the weak label \(\bm{w}\) is typically generated from instance-level labels \(\bm{y}\) through the aggregation rule \(\phi\). Since \(\bm{y}\) remains unobservable, it acts as a latent variable, which aligns with the classic framework of the Expectation Maximization (EM) algorithm
Therefore, we explore the relationship between our method and the EM algorithm, and find that optimization of the loss in Theorem \ref{thm: URE} is somewhat equivalent to the EM algorithm \cite{EM1, EM2}:

\begin{theorem}[Equivalent to EM algorithm]\label{thm: EM} 
    Let \(\theta=\{\theta_1\text{ (state evolution module)}, \theta_2 \text{ (backbone)}\}\). Then  optimizing \(\tilde{L}\) in Theorem \ref{thm: URE} with BCE under most weak supervisions is equivalent to applying EM (via minimization) or GEM (via gradient descent) algorithms:

    \begin{IEEEitemize}[\setlength{\IEEElabelindent}{1pt}\setlength{\itemindent}{1pt}\setlength{\labelsep}{3pt}]
        \item 
        E step \(\iff\) During the forward process, closing the gradient about \(\theta_1\) in \(P(\bm{Y}^k=\bm{e}_j|\bm{X}^{[K]}=\bm{x}^{[K]}, \bm{W}=\bm{w};\theta_1)\) when calculating URE .

        \item 
        M step \(\iff\) In the backward process, the optimization on \(\theta\) involves only the gradient in \(l(f_i(\bm{x}^k), \lambda;\theta)\).
    \end{IEEEitemize}
\end{theorem}
Specifically, the loss calculation process with estimation of latent probabilities akin to the E step, and the model optimization while neglecting the coefficient in the loss \(\tilde{L}\) resembles the M step. 
In contrast, GLWS \cite{GWSL_ICML_2024}, which also builds upon the EM framework, models the weak supervision through its NFA. However, we could demonstrate that the weakly supervised scenario representable by GLWS’s NFA can also be captured by our loopy Bayesian network. Therefore, our method retains the theoretical rigor of EM-based approaches while generalizing to a broader class of problems and expanding the scope of application scenarios.
Moreover, we also provide a theoretical error bound of our methods in the supplementary materials.




\section{Experiments}
\noindent
In this section, we demonstrate the effectiveness of FastBUS by comparing it with state-of-the-art approaches under various settings. Specifically, we report experimental results in four parts, including aggregation supervision, pairwise supervision, imperfect supervision, and incomplete supervision. We mainly present results on CIFAR-10 \cite{CIFAR_10_100_data}, CIFAR-100 \cite{CIFAR_10_100_data}, and STL-10 \cite{STL_10_data} in the main paper, while other datasets like MNIST \cite{MNIST_data} and F-MNIST \cite{F_MNIST_data} are provided in the appendices. In addition, we further conduct analytical experiments on training time and network structure, and supplement additional experiments in appendices such as the selectivity of rank \(r\). Moreover, our code integrates basic classifiers into the GBP training framework.

\begin{table}[!t]
\centering
\caption{Accuracy on Aggregation Supervision for Three Datasets. All Results Are Averaged over Five runs.}
\label{tab: experiment aggregation}
\begin{tabularx}{\hsize}{>{\raggedright\arraybackslash}p{1.45cm}||YY|YY|YY}
\toprule
    \rowcolor{tablegray}
    \textbf{Dataset}   
    & \multicolumn{2}{c|}{\textbf{CIFAR10}} 
    & \multicolumn{2}{c|}{\textbf{CIFAR100}} 
    & \multicolumn{2}{c}{\textbf{STL10}}  \\
    
\midrule
\rowcolor{tablegray}
    \multicolumn{7}{c}{\textbf{Multi-Instance Learning}}  \\
    
\midrule
    $N$         
    & $800$      
    & $1200$            
    & $1000$     
    & $2500$              
    & $500$      
    & $2000$             \\

    $K$          
    & \scalebox{0.85}{\(\mathcal{N}(8, 1)\)}  
    & \scalebox{0.85}{\(\mathcal{N}(5, 1)\)}     
    & \scalebox{0.85}{\(\mathcal{N}(10, 2)\)}
    & \scalebox{0.85}{\(\mathcal{N}(8, 1)\)}       
    & \scalebox{0.85}{\(\mathcal{N}(12, 2)\)} 
    & \scalebox{0.85}{\(\mathcal{N}(5, 1)\)}         \\
    
\midrule
    \textbf{Count Loss} 
    & $64.95~$ \tiny{$\pm 0.61$}  
    & $70.31~ $ \tiny{$\pm 0.83$}           
    & $29.50~ $  \tiny{$\pm 1.29$}  
    & $46.87$    \tiny{$\pm 0.82$}          
    & $49.84~$  \tiny{$\pm 2.62$}   
    & $73.05~$   \tiny{$\pm 1.37$}         \\
    
    \textbf{UUM}        
    & $14.90~$ \tiny{$\pm 0.62$}  
    & $16.23~$  \tiny{$\pm 0.66$}            
    & $1.50~$   \tiny{$\pm 0.62$}   
    & $1.58 $   \tiny{$\pm 0.51$}           
    & $16.95~$  \tiny{$\pm 2.84$}   
    & $14.45~$   \tiny{$\pm 2.26$}         \\
    
    \textbf{GLWS}        
    & $68.97~$ \tiny{$\pm 0.79$}    
    & $74.04~$   \tiny{$\pm 0.67$}          
    & $25.77~$  \tiny{$\pm 1.26$}  
    & $44.86 $   \tiny{$\pm 1.92$}           
    & $53.95~$   \tiny{$\pm 0.95 $}  
    & $76.28~$  \tiny{$\pm 1.31$}         \\

    \rowcolor{tableblue}
    \textbf{FastBUS}       
    & $\mathbf{71.02}$ \tiny{$\pm 0.81$}  
    & $\mathbf{74.56}$ \tiny{$\pm 0.63$}            
    & $\mathbf{34.74}$  \tiny{$\pm 1.50$}   
    & $\mathbf{48.50}$  \tiny{$\pm 1.52$}           
    & $\mathbf{55.49}$   \tiny{$\pm 0.16$}  
    & $\mathbf{76.80}$  \tiny{$\pm 1.28$}          \\
    
\midrule
    \rowcolor{tablegray}
    \multicolumn{7}{c}{\textbf{Learn from Label Proportion}} \\ 

\midrule
    \(N\)        
    & 1200     
    & 2000             
    & 500      
    & 1000              
    & 500     
    & 2000            \\

    \(K\)          
    & \scalebox{0.85}{\(\mathcal{N}(10, 2)\)}
    & \scalebox{0.85}{\(\mathcal{N}(12, 2)\)}        
    & \scalebox{0.85}{\(\mathcal{N}(5, 1)\)}  
    & \scalebox{0.85}{\(\mathcal{N}(12, 2)\)}          
    & \scalebox{0.85}{\(\mathcal{N}(5, 1)\)}  
    & \scalebox{0.85}{\(\mathcal{N}(5, 1)\)}         \\
    
\midrule
    \textbf{LLP VAT}   
    & $75.83$ \tiny{$\pm 0.63$}  
    & $81.37$  \tiny{$\pm 0.85$}       
    & $13.41~$  \tiny{$\pm 0.75$}  
    & $10.76~$   \tiny{$\pm 0.83$}           
    & $57.86~$  \tiny{$\pm 2.86$}   
    & $74.83~$  \tiny{$\pm 1.49$}          \\
    
    \textbf{Count Loss}
    & $79.14~$  \tiny{$\pm 0.49$}  
    & $84.13~$  \tiny{$\pm 0.59$}           
    & $18.12~$  \tiny{$\pm 0.69$}  
    & $35.84~$   \tiny{$\pm 0.62$}           
    & $61.45~$  \tiny{$\pm 1.42$}   
    & $77.55~$  \tiny{$\pm 1.26$}         \\
    
    \textbf{UUM}        
    & -        
    & -                
    & $16.97~$  \tiny{$\pm 1.26$}   
    & -                 
    & $58.94~$ \tiny{$\pm 0.78$}   
    & $76.38$ \tiny{$\pm 0.49$}                 \\
    
    \textbf{GLWS}        
    & $78.97~$  \tiny{$\pm 0.28$}  
    & $84.08~$  \tiny{$\pm 0.72$}         
    & $17.23~$  \tiny{$\pm 1.18$}   
    & $36.95~$   \tiny{$\pm 0.95$}          
    & $58.16~$  \tiny{$\pm 1.57$}   
    & $76.50~$  \tiny{$\pm 0.85$}          \\

    \rowcolor{tableblue}
    \textbf{FastBUS}       
    & $\mathbf{80.93}$ \tiny{$\pm 0.42$}  
    & $\mathbf{86.02}$  \tiny{$\pm 0.93$}           
    & $\mathbf{20.99}$  \tiny{$\pm 0.94$}   
    & $\mathbf{44.50}$  \tiny{$\pm 0.78$}          
    & $\mathbf{63.61}$  \tiny{$\pm 1.24$}   
    & $\mathbf{78.21}$  \tiny{$\pm 0.62$}        \\
\bottomrule
\end{tabularx}
\end{table}

\subsection{Aggregation Supervision}\noindent
\textbf{Setup.} 
First, we validate our method under aggregation supervision, including the \emph{MultiIns} and \emph{LProp} settings. Specifically, following GLWS \cite{GWSL_ICML_2024}, we randomly aggregate instances into bags, where the number of instances \(K\) per bag is sampled from a Gaussian distribution. Notably, this aggregation process produces a multi-label dataset. We compare our approach against established baselines such as GLWS \cite{GWSL_ICML_2024}, Count Loss \cite{GWSL_NIPS_2023}, and UUM \cite{GWSL_ICML_2023}. For \emph{LProp}, we additionally include LLP-VAT \cite{LLP_VAT} as a baseline. Detailed training hyperparameters are provided in the appendix.

\textbf{Results.} 
Table \ref{tab: experiment aggregation} presents the results for aggregation supervision. Our method significantly outperforms other approaches, achieving improvements of 2.05\% on CIFAR-10, 5.24\% on CIFAR-100, and 1.54\% on STL-10 for \emph{MultiIns}. For \emph{LProp}, our method surpasses baselines by 1.79\% on CIFAR-10, 3.76\% on CIFAR-100, and 2.16\% on STL-10. Beyond superior performance, our method also achieves the fastest training speed – both advancements can be attributed to our accelerated unified Bayesian Network. Specifically, methods like GLWS \cite{GWSL_ICML_2024} employ instance-specific hard-coded probability computation schemes for \emph{LProp}, which not only leads to slow runtime due to the limitation of batch processing but also fails to effectively learn data generation patterns, which our modular network successfully captures. Furthermore, UUM \cite{GWSL_ICML_2023} demonstrates inferior performance across all settings and only works reliably with small bag sizes in \emph{LProp}, as it relies on a simplistic direct computation method with exponential time complexity and limited capability for learning data patterns efficiently.

\subsection{Pairwise Supervision}

\begin{table}[!t]
\centering
\caption{Accuracy on Hard-Label Pairwise Supervision for Three datasets. All Results Are Averaged over Five Runs.}
\label{tab: experiment pairwise hard}
\begin{tabularx}{\hsize}{>{\raggedright\arraybackslash}m{1.15cm}||YY|YY|YY}

\toprule
    \rowcolor{tablegray}
    \textbf{Dataset}       
    & \multicolumn{2}{c|}{\textbf{CIFAR-10 }}                
    & \multicolumn{2}{c|}{\textbf{CIFAR-100 }}               
    & \multicolumn{2}{c}{\textbf{STL-10 }}                   \\

\midrule
    \rowcolor{tablegray}
    \multicolumn{7}{c}{\textbf{Pairwise Comparison}}  \\
    
\midrule
    \({N}\)              
    & $7500$                    
    & $10000$                   
    & $5000$                    
    & $7500$                   
    & $2000$                    
    & $6000$                     \\
    
    \textbf{Prior}         
    & $0.65$                    
    & $0.45$                    
    & $0.65$                    
    & $0.45$                    
    & $0.65$                    
    & $0.45$                     \\
    
\midrule
    \multirow{2}{*}{\parbox[t]{1.1cm}{\textbf{PcABS}}}   
    & $70.19$ \tiny{$ \pm 0.23$}         
    & $72.97$ \tiny{$ \pm 0.28$}          
    & $55.87$ \tiny{$ \pm 0.39$}          
    & $60.13$ \tiny{$ \pm 0.53$}          
    & $70.98$ \tiny{$ \pm 0.52$}          
    & $72.16$ \tiny{$ \pm 0.43$}           \\
    
    \multirow{2}{*}{\parbox[t]{1.1cm}{\textbf{PcReLU}}}     
    & $71.64$ \tiny{$ \pm 0.46$}         
    & $74.97$ \tiny{$ \pm 0.34$}          
    & $56.88$ \tiny{$ \pm 0.31$}          
    & $63.24$ \tiny{$ \pm 0.61$}          
    & $72.03$ \tiny{$ \pm 0.37$}          
    & $74.66$ \tiny{$ \pm 0.36$}           \\[1.1em]
    
    \multirow{2}{*}{\parbox[t]{1.1cm}{\textbf{PcTeacher}}} 
    & $60.29$ \tiny{$ \pm 0.35$}          
    & $60.62$ \tiny{$ \pm 0.56$}          
    & $57.97$ \tiny{$ \pm 0.47$}         
    & $59.06$ \tiny{$ \pm 0.31$}          
    & $70.21$ \tiny{$ \pm 0.23$}          
    & $70.36$ \tiny{$ \pm 0.48$}           \\[1.1em]
    
    \multirow{2}{*}{\parbox[t]{1.2cm}{\textbf{PcUnbias}}} 
    & $67.58$ \tiny{$ \pm 0.71$}          
    & $71.72$ \tiny{$ \pm 0.23$}          
    & $54.92$ \tiny{$ \pm 0.26$}          
    & $62.55$ \tiny{$ \pm 0.42$}          
    & $70.25$ \tiny{$ \pm 0.39$}          
    & $71.05$ \tiny{$ \pm 0.31$}           \\[1.1em]

    \multirow{2}{*}{\parbox[t]{1.1cm}{\textbf{Rank Pruning}}}   
    & $61.25$ \tiny{$ \pm 0.24$}          
    & $62.50$ \tiny{$ \pm 0.07$}          
    & $57.83$ \tiny{$ \pm 0.53$}          
    & $59.06$ \tiny{$ \pm 0.47$}          
    & $70.85$ \tiny{$ \pm 0.28$}          
    & $69.26$ \tiny{$ \pm 0.47$}           \\[1.1em]
    
    \multirow{2}{*}{\textbf{GLWS}}           
    & $72.30$ \tiny{$ \pm 0.19$}           
    & $75.08$ \tiny{$ \pm 0.18$}          
    & $61.22$ \tiny{$ \pm 0.22$}          
    & $62.57$ \tiny{$ \pm 0.13$}          
    & $71.40$ \tiny{$ \pm 0.31$}          
    & $76.67$ \tiny{$ \pm 0.25$}           \\[1.1em]

    \rowcolor{tableblue}
    \multirow{2}{*}{\textbf{FastBUS}}          
    & $\mathbf{73.72}$ \tiny{$ \pm 0.21$}
    & $\mathbf{75.39}$ \tiny{$ \pm 0.26$} 
    & $\mathbf{62.09}$ \tiny{$ \pm 0.15$} 
    & $\mathbf{64.45}$ \tiny{$ \pm 0.35$} 
    & $\mathbf{73.67}$ \tiny{$ \pm 0.24$} 
    & $\mathbf{78.48}$ \tiny{$ \pm 0.33$}  \\

\midrule 
    \rowcolor{tablegray}
    \multicolumn{7}{c}{\textbf{Pairwise Similarity}}   \\ 
    
\midrule                 
    \({N}\)           
    & $5000$                    
    & $10000$                   
    & $2000$                    
    & $6000$                    
    & $2500$                    
    & $5000$                     \\

    \textbf{Prior}          
    & $0.65$                   
    & $0.45$                    
    & $0.65$                    
    & $0.45$                    
    & $0.65$                    
    & $0.45$                     \\
    
\midrule
    \multirow{2}{*}{\textbf{RiskSD} }       
    & $63.28$ \tiny{$ \pm 0.61$}          
    & $61.37$ \tiny{$ \pm 0.49$}          
    & $56.18$ \tiny{$ \pm 0.76$}         
    & $61.22$ \tiny{$ \pm 0.75$}         
    & $67.05$ \tiny{$ \pm 0.41$}          
    & $59.89$ \tiny{$ \pm 0.40$}           \\
    
    \multirow{2}{*}{\textbf{UUM}}            
    & $65.33$ \tiny{$ \pm 0.47$}         
    & $67.87$ \tiny{$ \pm 0.37$}          
    & $59.19$ \tiny{$ \pm 0.61$}          
    & $59.72$ \tiny{$ \pm 0.52$}          
    & $74.96$ \tiny{$ \pm 0.35$}          
    & $73.63$ \tiny{$ \pm 0.16$}           \\
    
    \multirow{2}{*}{\textbf{GLWS}}           
    & $65.17$ \tiny{$ \pm 0.19$}          
    & $63.98$ \tiny{$ \pm 0.18$}          
    & $57.65$ \tiny{$ \pm 0.42$}          
    & $59.25$ \tiny{$ \pm 0.34$}          
    & $74.88$ \tiny{$ \pm 0.44$}          
    & $74.37$ \tiny{$ \pm 0.42$}           \\

    \rowcolor{tableblue}
    \multirow{2}{*}{\textbf{FastBUS}}           
    & $\mathbf{65.58}$ \tiny{$\pm 0.26$}      
    & $\mathbf{70.44}$ \tiny{$\pm 0.36$}        
    & $\mathbf{60.28}$ \tiny{$\pm 0.31$}          
    & $\mathbf{61.66}$ \tiny{$\pm 0.29$}         
    & $\mathbf{75.50}$ \tiny{$\pm 0.32$}         
    & $\mathbf{75.23}$ \tiny{$\pm 0.46$}          \\
\bottomrule
\end{tabularx}
\end{table}
\begin{table}[!t]
\centering
\caption{Accuracy on Soft-Label Pairwise Supervision for Three datasets. All Results Are Averaged over Five Runs.}
\label{tab: experiment pairwise soft}
\begin{tabularx}{\hsize}{>{\raggedright\arraybackslash}p{1.3cm}||YY|YY|YY}

\toprule
    \rowcolor{tablegray}
    \textbf{Dataset}       
    & \multicolumn{2}{c|}{\textbf{CIFAR-10 }}                
    & \multicolumn{2}{c|}{\textbf{CIFAR-100 }}               
    & \multicolumn{2}{c}{\textbf{STL-10 }}                   \\

\midrule
    \rowcolor{tablegray}
    \multicolumn{7}{c}{\textbf{Similarity Confidence}}  \\
    
\midrule
    \({N}\)              
    & $4000$                    
    & $10000$                   
    & $3000$                    
    & $10000$                    
    & $3000$                    
    & $10000$                     \\
    
    \textbf{Prior}         
    & $0.45$      
    & $0.65$                    
    & $0.45$                    
    & $0.65$                    
    & $0.45$                    
    & $0.65$                     \\
    
\midrule
    \multirow{2}{*}{\parbox[t]{1.1cm}{\textbf{ScABS}}}   
    & $61.26$ \tiny{$ \pm 0.32$}         
    & $68.68$ \tiny{$ \pm 0.51$}          
    & $60.52$ \tiny{$ \pm 0.21$}              
    & $60.02$ \tiny{$ \pm 0.47$}          
    & $66.08$ \tiny{$ \pm 0.17$}          
    & $71.32$ \tiny{$ \pm 0.29$}           \\
    
    \multirow{2}{*}{\parbox[t]{1.1cm}{\textbf{SCReLU}}}     
    & $60.21$ \tiny{$ \pm 0.51$}         
    & $68.51$ \tiny{$ \pm 0.43$}          
    & $60.45$ \tiny{$ \pm 0.17$}          
    & $60.04$ \tiny{$ \pm 0.38$}          
    & $68.02$ \tiny{$ \pm 0.26$}          
    & $70.98$ \tiny{$ \pm 0.19$}           \\
    
    \multirow{2}{*}{\parbox[t]{1.1cm}{\textbf{ScNNABS}}} 
    & $61.11$ \tiny{$ \pm 0.46$}          
    & $68.50$ \tiny{$ \pm 0.72$}          
    & $60.46$ \tiny{$ \pm 0.16$}         
    & $60.02$ \tiny{$ \pm 0.52$}          
    & $69.41$ \tiny{$ \pm 0.32$}          
    & $71.06$ \tiny{$ \pm 0.58$}           \\

    \multirow{2}{*}{\parbox[t]{1.1cm}{\textbf{ScUnbias}}}   
    & $60.47$ \tiny{$ \pm 0.37$}          
    & $67.75$ \tiny{$ \pm 0.38$}          
    & $60.40$ \tiny{$ \pm 0.01$}          
    & $60.03$ \tiny{$ \pm 0.43$}          
    & $69.71$ \tiny{$ \pm 0.31$}          
    & $70.72$ \tiny{$ \pm 0.41$}           \\
    
    \multirow{2}{*}{\textbf{GLWS}}           
    & $65.17$ \tiny{$ \pm 0.45$}           
    & $71.17$ \tiny{$ \pm 0.79$}          
    & $61.53$ \tiny{$ \pm 0.34$}          
    & $61.10$ \tiny{$ \pm 0.25$}          
    & $68.72$ \tiny{$ \pm 0.53$}          
    & $67.82$ \tiny{$ \pm 0.37$}           \\

    \rowcolor{tableblue}
    \multirow{2}{*}{\textbf{FastBUS}}          
    & $\mathbf{65.39}$ \tiny{$ \pm 0.35$}
    & $\mathbf{72.29}$ \tiny{$ \pm 0.64$} 
    & $\mathbf{62.44}$ \tiny{$ \pm 0.05$} 
    & $\mathbf{61.36}$ \tiny{$ \pm 0.18$} 
    & $\mathbf{68.95}$ \tiny{$ \pm 0.34$} 
    & $\mathbf{71.49}$ \tiny{$ \pm 0.24$}  \\

\midrule 
    \rowcolor{tablegray}
    \multicolumn{7}{c}{\textbf{Confidence Difference}}   \\ 
    
\midrule                 
    \({N}\)           
    & $3000$                    
    & $10000$                   
    & $4000$                    
    & $10000$                    
    & $6000$                    
    & $10000$                     \\

    \textbf{Prior}          
    & $0.45$                   
    & $0.65$                    
    & $0.45$                    
    & $0.65$                    
    & $0.45$                    
    & $0.65$                     \\
    
\midrule
    \multirow{2}{*}{\textbf{CdABS} }       
    & $70.33$ \tiny{$ \pm 0.35$}          
    & $\mathbf{76.85}$ \tiny{$ \pm 0.19$}          
    & $62.98$ \tiny{$ \pm 0.76$}         
    & $66.41$ \tiny{$ \pm 0.45$}         
    & $69.15$ \tiny{$ \pm 0.32$}          
    & $72.97$ \tiny{$ \pm 0.28$}           \\
    
    \multirow{2}{*}{\textbf{CdReLU}}            
    & $70.17$ \tiny{$ \pm 0.30$}         
    & $76.48$ \tiny{$ \pm 0.47$}          
    & $63.13$ \tiny{$ \pm 0.51$}          
    & $66.41$ \tiny{$ \pm 0.62$}          
    & $69.03$ \tiny{$ \pm 0.13$}          
    & $72.05$ \tiny{$ \pm 0.19$}           \\

    \multirow{2}{*}{\textbf{CdUnbias}}            
    & $69.36$ \tiny{$ \pm 0.37$}         
    & $74.96$ \tiny{$ \pm 0.58$}          
    & $\mathbf{63.17}$ \tiny{$ \pm 0.39$}          
    & $66.37$ \tiny{$ \pm 0.47$}          
    & $68.87$ \tiny{$ \pm 0.26$}          
    & $72.48$ \tiny{$ \pm 0.09$}           \\
    
    \multirow{2}{*}{\textbf{GLWS}}           
    & $69.71$ \tiny{$ \pm 0.19$}          
    & $76.04$ \tiny{$ \pm 0.43$}          
    & $60.95$ \tiny{$ \pm 0.20$}          
    & $59.55$ \tiny{$ \pm 0.48$}          
    & $68.81$ \tiny{$ \pm 0.27$}          
    & $72.57$ \tiny{$ \pm 0.52$}           \\

    \rowcolor{tableblue}
    \multirow{2}{*}{\textbf{FastBUS}}           
    & $\mathbf{70.56}$ \tiny{$\pm 0.21$}      
    & $76.17$ \tiny{$\pm 0.18$}        
    & $60.28$ \tiny{$\pm 0.43$}          
    & $\mathbf{66.73}$ \tiny{$\pm 0.37$}         
    & $\mathbf{71.12}$ \tiny{$\pm 0.25$}         
    & $\mathbf{74.32}$ \tiny{$\pm 0.38$}          \\
\bottomrule
\end{tabularx}
\end{table}

\begin{table}[!t]
\centering
\caption{Accuracy on Imperfect Supervision for Three datasets. All Results Are Averaged over Five Runs. }
\label{tab: experiment imperfect}
\begin{tabularx}{\hsize}{>{\raggedright\arraybackslash}p{1.25cm}||YY|YY|YY}

\toprule
    \rowcolor{tablegray}
    \textbf{Dataset}       
    & \multicolumn{2}{c|}{\textbf{CIFAR-10 }}                
    & \multicolumn{2}{c|}{\textbf{CIFAR-100 }}               
    & \multicolumn{2}{c}{\textbf{STL-10 }}                   \\

\midrule
    \rowcolor{tablegray}
    \multicolumn{7}{c}{\textbf{Partial Label}}  \\
    
\midrule
    \({N}\)              
    & \multicolumn{2}{c|}{20000}                   
    & \multicolumn{2}{c|}{20000}                        
    & \multicolumn{2}{c}{20000}    \\
    
    \textbf{Ratio}         
    & $0.30$               $ $     
    & $0.70$                    
    & $0.03$                    
    & $0.10$                    
    & $0.05$                    
    & $0.50$                     \\
    
\midrule
    \multirow{2}{*}{\parbox[t]{1.2cm}{\textbf{CC}}}   
    & $91.57$ \tiny{$ \pm 0.15$}         
    & $83.58$ \tiny{$ \pm 0.36$}          
    & $56.47$ \tiny{$ \pm 0.24$}          
    & $53.74$ \tiny{$ \pm 0.18$}          
    & $85.72$ \tiny{$ \pm 0.45$}          
    & $82.05$ \tiny{$ \pm 0.72$}           \\
    
    
    \multirow{2}{*}{\parbox[t]{1.2cm}{\textbf{PRODEN}}} 
    & $91.86$ \tiny{$ \pm 0.21$}          
    & $85.97$ \tiny{$ \pm 0.35$}          
    & $57.19$ \tiny{$ \pm 0.16$}         
    & $53.73$ \tiny{$ \pm 0.42$}          
    & $85.88$ \tiny{$ \pm 0.39$}          
    & $82.49$ \tiny{$ \pm 0.73$}           \\
    
    \multirow{2}{*}{\parbox[t]{1.2cm}{\textbf{PiCO}}} 
    & $92.49$ \tiny{$ \pm 0.48$}          
    & $87.37$ \tiny{$ \pm 0.53$}          
    & $57.28$ \tiny{$ \pm 0.39$}          
    & $55.07$ \tiny{$ \pm 0.44$}          
    & $86.07$ \tiny{$ \pm 0.64$}          
    & $82.85$ \tiny{$ \pm 0.47$}           \\

    \multirow{2}{*}{\parbox[t]{1.2cm}{\textbf{RCR}}}   
    & $92.73$ \tiny{$ \pm 0.08$}          
    & $87.53$ \tiny{$ \pm 0.21$}          
    & $57.74$ \tiny{$ \pm 0.14$}          
    & $54.83$ \tiny{$ \pm 0.19$}          
    & $86.14$ \tiny{$ \pm 0.42$}          
    & $83.17$ \tiny{$ \pm 0.58$}           \\
    
    \multirow{2}{*}{\textbf{GLWS}}           
    & $92.98$ \tiny{$ \pm 0.09$}           
    & $87.86$ \tiny{$ \pm 0.15$}          
    & $\mathbf{58.42}$ \tiny{$ \pm 0.28$}          
    & $55.26$ \tiny{$ \pm 0.37$}          
    & $85.93$ \tiny{$ \pm 0.31$}          
    & $83.28$ \tiny{$ \pm 0.46$}           \\

    \rowcolor{tableblue}
    \multirow{2}{*}{\textbf{FastBUS}}          
    & $\mathbf{93.76}$ \tiny{$ \pm 0.25$}
    & $\mathbf{88.29}$ \tiny{$ \pm 0.32$} 
    & $57.96$ \tiny{$ \pm 0.13$} 
    & $\mathbf{56.44}$ \tiny{$ \pm 0.21$} 
    & $\mathbf{86.46}$ \tiny{$ \pm 0.29$} 
    & $\mathbf{83.46}$ \tiny{$ \pm 0.36$}  \\

\midrule 
    \rowcolor{tablegray}
    \multicolumn{7}{c}{\textbf{Noisy Label}}   \\ 
    
\midrule                 

    \textbf{Ratio}          
    & $0.3$                   
    & $0.7$                    
    & $0.3$                    
    & $0.7$                    
    & $0.3$                    
    & $0.7$                     \\
    
\midrule
    \multirow{2}{*}{\textbf{DivideMix} }       
    & $92.52$ \tiny{$ \pm 0.25$}          
    & $90.75$ \tiny{$ \pm 0.32$}          
    & $73.28$ \tiny{$ \pm 0.34$}         
    & $65.89$ \tiny{$ \pm 0.48$}         
    & $70.43$ \tiny{$ \pm 0.37$}          
    & $53.72$ \tiny{$ \pm 0.51$}           \\
    
    \multirow{2}{*}{\textbf{ELR}}            
    & $92.73$ \tiny{$ \pm 0.32$}         
    & $91.28$ \tiny{$ \pm 0.35$}          
    & $72.83$ \tiny{$ \pm 0.27$}          
    & $65.76$ \tiny{$ \pm 0.41$}          
    & $70.92$ \tiny{$ \pm 0.29$}          
    & $53.94$ \tiny{$ \pm 0.46$}           \\

    \multirow{2}{*}{\textbf{SOP}}            
    & $93.15$ \tiny{$ \pm 0.34$}         
    & $91.14$ \tiny{$ \pm 0.36$}          
    & $\mathbf{73.67}$ \tiny{$ \pm 0.30$}          
    & $\mathbf{66.47}$ \tiny{$ \pm 0.49$}          
    & $72.61$ \tiny{$ \pm 0.29$}          
    & $54.47$ \tiny{$ \pm 0.58$}           \\

    \rowcolor{tableblue}
    \multirow{2}{*}{\textbf{FastBUS}}           
    & $\mathbf{93.58}$ \tiny{$\pm 0.18$}      
    & $\mathbf{92.24}$ \tiny{$\pm 0.29$}        
    & $72.91$ \tiny{$\pm 0.31$}          
    & $66.29$ \tiny{$\pm 0.46$}         
    & $\mathbf{72.91}$ \tiny{$\pm 0.28$}         
    & $\mathbf{54.66}$ \tiny{$\pm 0.45$}          \\
\bottomrule
\end{tabularx}
\end{table}

\noindent
\textbf{Setup.} 
We present experimental results for pairwise supervision under the previously baselines from GLWS \cite{GWSL_ICML_2024}, including PairComp (Pc) with its variants \cite{paircomp2021} and rank pruning \cite{rank_pruning} as baselines for \emph{PairComp}; RiskRD \cite{simunl2021} and UUM \cite{GWSL_ICML_2023} for \emph{PairSim}; SimConf (Sc) \cite{simconf2021} with variants for \emph{SimConf}; and ConfDiff (Cd) \cite{confdiff2023} with variants for \emph{ConfDiff}. Additionally, GLWS itself serves as a baseline. For soft-label pairwise supervision, we employ CLIP \cite{CLIP2021} as the confidence scoring model, consistent with GLWS \cite{GWSL_ICML_2024}. Aligned with prior works \cite{GWSL_ICML_2023, GWSL_ICML_2024, GWSL_NIPS_2020}, since classifiers in an identifiable manner cannot be directly learned from pairwise supervision data, we utilize the Hungarian algorithm \cite{kuhn1955hungarian} to determine the optimal prediction mapping.

\textbf{Results.} 
 Tables~\ref{tab: experiment pairwise hard} and~\ref{tab: experiment pairwise soft} summarize the experimental results under hard-label and soft-label pairwise supervision settings, respectively. For both \emph{PairComp} and \emph{PairSim}, our proposed method consistently surpasses all baseline approaches. Specifically, \emph{PairComp} demonstrates robust performance gains, achieving nearly a 1\% accuracy improvement across most evaluation scenarios. The advantage is even more pronounced for \emph{PairSim}, which attains a significant 2.57\% performance boost on the CIFAR-10 benchmark. In the soft-label supervision setting, our method maintains its superiority, delivering state-of-the-art results in the majority of cases. A particularly notable improvement is observed with \emph{ConfDiff} on the STL-10 dataset, where our approach yields a 1.97\% increase in accuracy. Furthermore, compared to GLWS~\cite{GWSL_ICML_2023}, our method shares the same advantage of not requiring any prior knowledge of class distributions, making it more flexible in real-world applications. Additionally, our approach achieves comparable computational efficiency, ensuring fast execution without sacrificing performance. 

\begin{table}[!t]
\centering
\caption{Accuracy on Incomplete Supervision for Three datasets. All Results Are Averaged over Five Runs.}
\label{tab: experiment incomplete}
\begin{tabularx}{\hsize}{>{\raggedright\arraybackslash}p{1.65cm}||YY|YY|YY}

\toprule
    \rowcolor{tablegray}
    \textbf{Dataset}       
    & \multicolumn{2}{c|}{\textbf{CIFAR-10 }}                
    & \multicolumn{2}{c|}{\textbf{CIFAR-100 }}               
    & \multicolumn{2}{c}{\textbf{STL-10 }}                   \\

\midrule
    \rowcolor{tablegray}
    \multicolumn{7}{c}{\textbf{Positive Unlabeled}}  \\
    
\midrule
    \({N_{pos}}\)              
    & $250$                    
    & $800$                   
    & $1500$                    
    & $3000$                    
    & $250$                    
    & $800$                     \\

\midrule
    \multirow{2}{*}{\parbox[t]{1.1cm}{\textbf{CVIR}}}   
    & $87.76$ \tiny{$ \pm 1.15$}         
    & $91.19$ \tiny{$ \pm 0.94$}          
    & $80.63$ \tiny{$ \pm 0.57$}              
    & $\mathbf{84.25}$ \tiny{$ \pm 0.34$}          
    & $71.76$ \tiny{$ \pm 1.48$}          
    & $76.90$ \tiny{$ \pm 1.17$}           \\
    
    \multirow{2}{*}{\parbox[t]{1.5cm}{\textbf{DIST PU}}}     
    & $81.67$ \tiny{$ \pm 2.87$}         
    & $83.94$ \tiny{$ \pm 1.69$}          
    & $74.68$ \tiny{$ \pm 1.21$}          
    & $77.19$ \tiny{$ \pm 0.85$}          
    & $72.87$ \tiny{$ \pm 1.46$}          
    & $74.64$ \tiny{$ \pm 0.74$}           \\
    
    \multirow{2}{*}{\parbox[t]{1.1cm}{\textbf{NN PU}}} 
    & $\mathbf{88.12}$ \tiny{$ \pm 0.94$}          
    & $89.86$ \tiny{$ \pm 0.68$}          
    & $76.22$ \tiny{$ \pm 0.89$}         
    & $77.96$ \tiny{$ \pm 0.51$}          
    & $71.23$ \tiny{$ \pm 0.72$}          
    & $79.46$ \tiny{$ \pm 0.37$}           \\
    
    \multirow{2}{*}{\parbox[t]{1.1cm}{\textbf{U PU}}} 
    & $73.75$ \tiny{$ \pm 1.82$}          
    & $86.24$ \tiny{$ \pm 0.83$}          
    & $67.64$ \tiny{$ \pm 0.76$}          
    & $72.14$ \tiny{$ \pm 0.69$}          
    & $71.52$ \tiny{$ \pm 0.82$}          
    & $75.35$ \tiny{$ \pm 0.55$}           \\

    \multirow{2}{*}{\parbox[t]{1.1cm}{\textbf{Var PU}}}   
    & $69.12$ \tiny{$ \pm 1.43$}          
    & $88.11$ \tiny{$ \pm 0.95$}          
    & $73.05$ \tiny{$ \pm 0.83$}          
    & $78.11$ \tiny{$ \pm 0.47$}          
    & $60.71$ \tiny{$ \pm 0.86$}          
    & $61.38$ \tiny{$ \pm 0.18$}           \\
    
    \multirow{2}{*}{\textbf{Count Loss}}           
    & $79.36$ \tiny{$ \pm 0.72$}           
    & $82.57$ \tiny{$ \pm 0.46$}          
    & $71.59$ \tiny{$ \pm 1.12$}          
    & $74.36$ \tiny{$ \pm 0.83$}          
    & $73.85$ \tiny{$ \pm 0.43$}          
    & $76.49$ \tiny{$ \pm 0.57$}           \\
    
    \multirow{2}{*}{\textbf{GLWS}}           
    & $87.85$ \tiny{$ \pm 0.39$}           
    & $\mathbf{91.81}$ \tiny{$ \pm 0.26$}          
    & $80.76$ \tiny{$ \pm 0.45$}          
    & $83.59$ \tiny{$ \pm 0.18$}          
    & $76.77$ \tiny{$ \pm 0.47$}          
    & $80.78$ \tiny{$ \pm 0.24$}           \\

    \rowcolor{tableblue}
    \multirow{2}{*}{\textbf{FastBUS}}        	  
    & $87.51$ \tiny{$ \pm 0.17$}
    & $91.56$ \tiny{$ \pm 0.25$} 
    & $\mathbf{80.93}$ \tiny{$ \pm 0.34$} 
    & $83.68$ \tiny{$ \pm 0.26$} 
    & $\mathbf{78.21}$ \tiny{$ \pm 0.42$} 
    & $\mathbf{81.36}$ \tiny{$ \pm 0.29$}  \\

\midrule 
    \rowcolor{tablegray}
    \multicolumn{7}{c}{\textbf{Semi-Supervised}}   \\ 
    
\midrule                 
    \({N}_{labels}\)           
    & $250$                    
    & $800$                   
    & $250$                    
    & $800$                    
    & $250$                    
    & $800$                     \\

\midrule
    \multirow{2}{*}{\textbf{Pseudo Label} }       
    & $54.93$ \tiny{$ \pm 1.72$}          
    & $66.92$ \tiny{$ \pm 0.58$}          
    & $14.39$ \tiny{$ \pm 0.72$}         
    & $38.82$ \tiny{$ \pm 0.49$}         
    & $60.37$ \tiny{$ \pm 0.63$}          
    & $62.38$ \tiny{$ \pm 0.47$}           \\
    
    \multirow{2}{*}{\textbf{VAT}}            
    & $59.86$ \tiny{$ \pm 0.86$}         
    & $68.73$ \tiny{$ \pm 1.36$}          
    & $13.72$ \tiny{$ \pm 0.58$}          
    & $36.30$ \tiny{$ \pm 0.60$}          
    & $59.39$ \tiny{$ \pm 0.82$}          
    & $61.48$ \tiny{$ \pm 0.52$}           \\

    \multirow{2}{*}{\textbf{Mean Teacher}}            
    & $62.47$ \tiny{$ \pm 1.62$}         
    & $73.48$ \tiny{$ \pm 1.93$}          
    & $15.85$ \tiny{$ \pm 0.92$}          
    & $39.65$ \tiny{$ \pm 0.73$}          
    & $62.49$ \tiny{$ \pm 0.47$}          
    & $65.39$ \tiny{$ \pm 0.45$}           \\
    
    \multirow{2}{*}{\textbf{MixMatch}}            
    & $86.44$ \tiny{$ \pm 0.58$}         
    & $90.56$ \tiny{$ \pm 0.46$}          
    & $22.75$ \tiny{$ \pm 0.65$}          
    & $50.62$ \tiny{$ \pm 0.63$}          
    & $73.36$ \tiny{$ \pm 0.49$}          
    & $78.83$ \tiny{$ \pm 0.46$}           \\

    \rowcolor{tableblue}
    \multirow{2}{*}{\textbf{FastBUS}}           
    & $\mathbf{87.27}$ \tiny{$\pm 0.26$}      
    & $\mathbf{90.70}$ \tiny{$\pm 0.31$}        
    & $\mathbf{24.29}$ \tiny{$\pm 0.47$}          
    & $\mathbf{52.78}$ \tiny{$\pm 0.50$}         
    & $\mathbf{81.78}$ \tiny{$\pm 0.51$}         
    & $\mathbf{87.08}$ \tiny{$\pm 0.39$}          \\
\bottomrule
\end{tabularx}
\end{table}
 
\subsection{Imperfect Supervision} 
\noindent
\textbf{Setup.} For imperfect supervision, we conducted experiments under two settings: Partial Label and Noisy Label Learning. In the Partial Label learning, candidate labels are randomly introduced for each class according to a specified partial ratio, aligning with established methodologies~\cite{partial2022ICML, PartialLv, GWSL_ICML_2024}. For the Noisy Label learning, following common synthesis practices \cite{noisy2022ICML}, a portion of the ground-truth labels (determined by the noise ratio) is randomly replaced with incorrect categories. The baseline methods employed include CC \cite{partial2020NIPS}, 
PRODEN \cite{PartialLv}, PiCO \cite{PICO}, RCR \cite{partial2022ICML}, and GLWS \cite{GWSL_ICML_2024} for Partial Label learning, and DivideMix \cite{noisy2020ICLR}, ELR \cite{noisy2020NIPS}, and SOP \cite{noisy2022ICML} for Noisy Label learning. More experimental details can be found in the appendices.

\textbf{Results.} 
Table \ref{tab: experiment imperfect} presents the experimental results under the imperfect supervision scenario, encompassing two distinct settings: Partial Label and Noisy Label. 
Our method delivers competitive performance across diverse partially labeled datasets. On CIFAR-10, it achieves a 0.5\% accuracy improvement over existing approaches. For challenging high-partial-ratio scenarios on CIFAR-100, the approach demonstrates stronger robustness with a 1.2\% performance advantage compared to baselines. It also establishes state-of-the-art results on the STL-10 dataset under partial supervision. In noisy label learning, our method yields significant gains exceeding 1\% accuracy on both CIFAR-10, while maintaining competitive performance on CIFAR-100 and STL-10.

\subsection{Incomplete Supervision}
\noindent \textbf{Setup.}
For incomplete supervisions, we first examine the Positive-Unlabeled setting where \(N_{pos}\) samples are positively labeled with known class priors on unlabeled data. Following GLWS \cite{GWSL_ICML_2024}, we focus on binary classification by mapping selected categories to positive/negative classes. Baseline comparisons include CVIR \cite{CVIR}, Dist PU \cite{DISTPU}, NN PU \cite{posunl2017}, U PU \cite{posunl2017}, Var PU \cite{VarPU}, Count Loss \cite{GWSL_NIPS_2023}, and GLWS \cite{GWSL_ICML_2024}. The second scenario involves Semi-Supervised Learning with \(N_{labels}\) correctly labeled samples in multi-class classification, using Pseudo Label \cite{semisup_pseudo_label}, VAT \cite{semisup_VAT}, Mean Teacher \cite{semisup_mean_teacher}, and MixMatch \cite{semisup_MixMatch} as baselines.

\textbf{Result.}
As shown in Table~\ref{tab: experiment incomplete}, our method demonstrates superior performance on half of the Positive-Unlabeled benchmarks. This advantage is particularly noteworthy given the computational efficiency derived from our probability calculation mechanism, which shares conceptual similarities with \emph{LProp} (see appendices for more details). In Semi-Supervised Learning, consistent improvements are observed across multiple datasets: an 8.4\% accuracy gain on STL-10, competitive results on CIFAR-100 (within 1.5\% of top baselines on 250 labeled data), and SOTA performance on CIFAR-10. These outcomes highlight our method's balanced effectiveness across diverse incomplete supervisions.

\begin{figure*}
    \centering
    \includegraphics[width=0.9\linewidth]{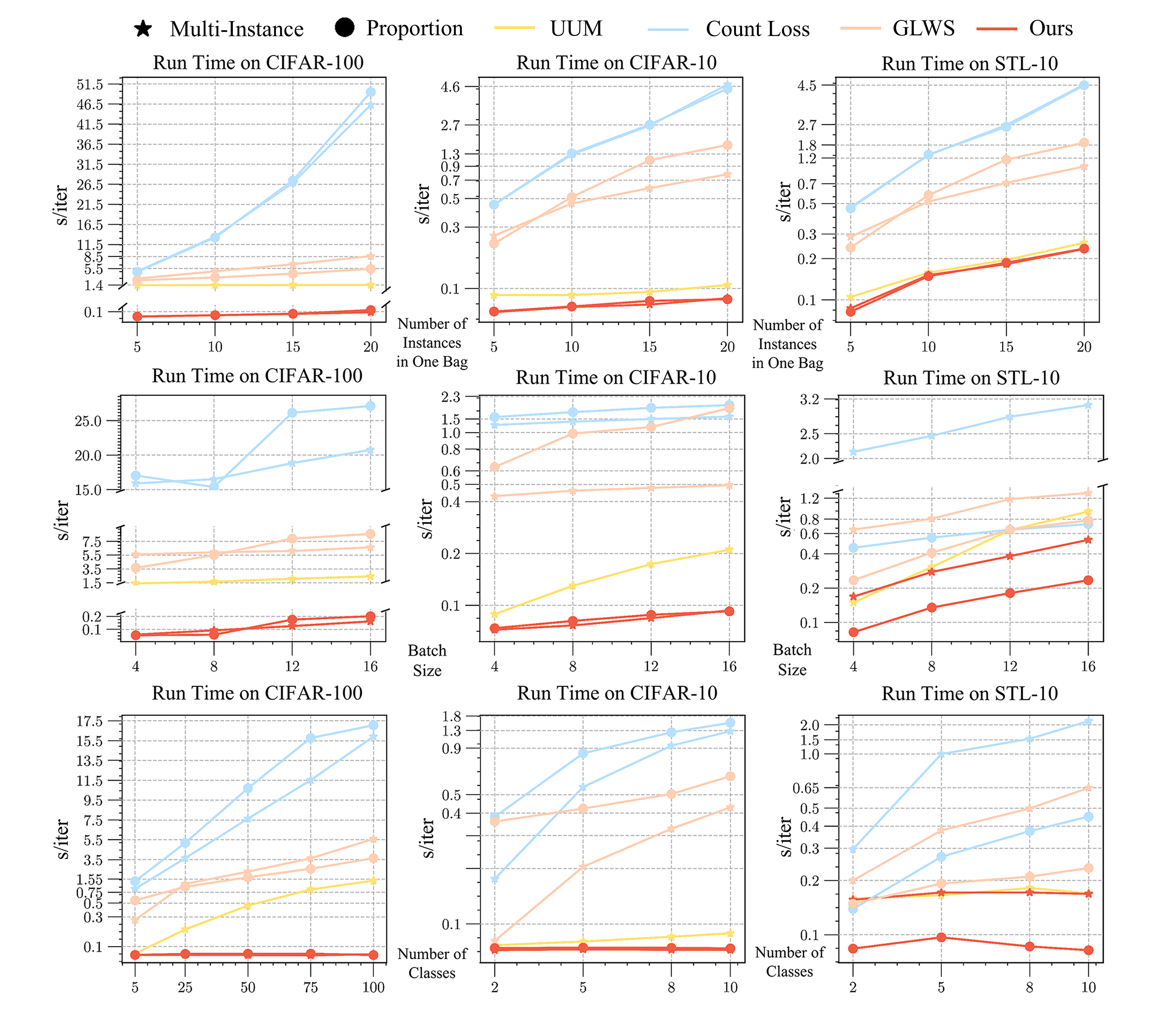}
    \caption{
    Runtime under varying instances per bag, batch sizes, and class counts. Results show our method maintains stable runtime on CIFAR-10/100 across these parameters, while others slow down. This efficiency arises from our unified Bayesian network, accelerated inference, and batch-level state evolution module, supporting a simultaneous multi-class batch process. In specific settings, we achieve up to \textbf{400x speedup} against Count Loss and GLWS.
    }
    \label{fig: run_time_all}
\end{figure*}

\begin{table}[!t]
\centering
\caption{Ablation Experiments using Different Modules to Learn Matrix \(\bm{U}\). Report Accuracy after Running All Results over Five Times.}
\label{tab: MLP ablation}
\begin{tabular}{@{}>{\raggedright\arraybackslash}p{1.05cm}||
                >{\centering\arraybackslash}m{0.83cm}
                >{\centering\arraybackslash}m{0.55cm}
                >{\centering\arraybackslash}m{1.cm}
                >{\centering\arraybackslash}m{0.83cm}
            >{\centering\arraybackslash}m{0.88cm}
            >{\centering\arraybackslash}m{1.1cm}@{}}
\toprule
    \textbf{Settings}       
    & \textbf{\emph{MultiIns}}               
    & \textbf{\emph{LProp}}              
    & \textbf{\emph{PairComp}} 
    & \textbf{\emph{PairSim}}
    & \textbf{\emph{SimConf}} 
    & \textbf{\emph{ConfDiff}}                \\

\midrule
    \textbf{MLP}               
    & $72.65$                    
    & $84.71$                   
    & $70.53$                    
    & $69.38$                    
    & $68.33$                    
    & $60.25$                     \\[0.5em]

    \parbox[c]{1.1cm}{\textbf{Ours w/o att.}}
    & $72.29$ 
    & $85.16$
    & $71.56$
    & $69.03$
    & $71.56$
    & $64.84$                     \\[1.em]

    \parbox[c]{1.1cm}{\textbf{Ours w/ ReLU}}
    & $15.98$ 
    & $84.93$
    & $60.18$
    & $70.06$
    & $60.08$
    & $60.27$                     \\[1.em]

    \parbox[c]{1.1cm}{\textbf{Ours w/ GELU}}
    & $73.37$ 
    & $84.81$
    & $60.83$
    & $70.34$
    & $60.75$
    & $60.74$                     \\[1.em]

    \textbf{Ours}         
    & $74.58$      
    & $86.14$                    
    & $73.81$                    
    & $70.57$                    
    & $72.31$                    
    & $70.65$                     \\
\bottomrule
\end{tabular}
\end{table}

\subsection{Empirical Analysis}
\noindent \textbf{RunTime.} 
Under \emph{LProp} and \emph{MultiIns}, Figure \ref{fig: run_time_all} compares the runtime of four general weakly-supervised methods across three datasets, with each row illustrating execution times under varying bag lengths (number of instances in each bag), batch sizes, and class numbers. Crucially, our method maintains stable runtime regardless of parameter changes. While UUM \cite{GWSL_ICML_2023} shows minor sensitivity to bag length and class count, Table \ref{tab: experiment aggregation} reveals its significant performance gap, and it fails in \emph{LProp} scenarios due to exponential complexity. In contrast, GLWS \cite{GWSL_ICML_2024} and Count Loss \cite{GWSL_NIPS_2023} exhibit substantial runtime increases as parameters scale—under \emph{MultiIns}, GLWS increases from 0.268s to 5.593s per iteration (20.87×) while Count Loss jumps from 0.848s to 15.911s (18.76×). This stems from GLWS’s sample/class-dependent NFA calculations (similar to the DFS tree specified by weak labels in figure \ref{fig: DFS}) and Count Loss’s quadratic complexity.

FastBUS achieves efficiency through three innovations: a unified Bayesian framework across various weak supervisions; low-rank optimization for single-step probability computation; and end-to-end learning of batch transition matrices. These collectively unify calculations across classes, reduce time complexity, and enable batch processing. Consequently, we achieve the fastest runtime: On CIFAR-100 (\emph{LProp}), our method operates at 0.11s (vs. Count Loss’s 49.52s, \textbf{450× faster}) when the bag length is 20, 0.2s (vs. Count Loss’s 27.12s, \textbf{135.6× faster}) as the batch size is 16, and 0.073s (vs. Count Loss’s 17.03s, \textbf{233.3× faster}) with 100 classes. Speedups range from \textbf{21.1–54.2×} on CIFAR-10 and \textbf{3.1-19.0×} on STL-10, demonstrating superior efficiency across datasets of varying scales and training configurations.

\textbf{Ablation of State Evolution Module.}
Table \ref{tab: MLP ablation} compares the results of learning the transition matrix \(\bm{U}\) using different modules, including standard MLP, our proposed State Evolution Module (SEM), SEM without attention, and SEM with other activation functions. The results show that compared to MLP, SEM can significantly improve the learning effect of the transition matrix, thereby enhancing the overall accuracy. The SEM performance decreased after removing the attention mechanism, especially in \emph{ConfDiff}, where it decreased by 5.81\%. Using other activation functions will reduce model performance, for example, using ReLU significantly reduces performance by 58.6\% in \emph{MultiIns}.

\textbf{Extension.} 
In the appendix, we have supplemented additional experiments, including results of various weak supervisions on other datasets, runtime performance under more settings, more ablation studies, and analysis experiments for the value of rank \(r\). It is worth noting that beyond the weakly-supervised scenarios discussed in the main paper, our method can readily adapt to combined weakly-supervised conditions, such as incomplete datasets with noisy labels or incompletely annotated pairwise datasets. This adaptability is achieved by repeatedly invoking the probabilistic computation methods on Bayesian networks.

\section{Conclusion}
To address limitations in prior studies, 
this paper proposes a unified Bayesian network framework. By employing the Generalized Belief Propagation algorithm over this structure, we achieve efficient estimation of latent probabilities across diverse weakly-supervised scenarios
. Furthermore, we introduce an optimized acceleration algorithm.
Extensive experiments across diverse scenarios and settings demonstrate that our method not only delivers superior performance but also achieves substantial speed improvements. We hope this work will inspire further research into universal weakly supervised learning frameworks.

\bibliographystyle{IEEEtran.bst}
\bibliography{IEEEabrv, main}

\begin{thebibliography}{10}
\providecommand{\url}[1]{#1}
\csname url@samestyle\endcsname
\providecommand{\newblock}{\relax}
\providecommand{\bibinfo}[2]{#2}
\providecommand{\BIBentrySTDinterwordspacing}{\spaceskip=0pt\relax}
\providecommand{\BIBentryALTinterwordstretchfactor}{4}
\providecommand{\BIBentryALTinterwordspacing}{\spaceskip=\fontdimen2\font plus
\BIBentryALTinterwordstretchfactor\fontdimen3\font minus \fontdimen4\font\relax}
\providecommand{\BIBforeignlanguage}[2]{{%
\expandafter\ifx\csname l@#1\endcsname\relax
\typeout{** WARNING: IEEEtran.bst: No hyphenation pattern has been}%
\typeout{** loaded for the language `#1'. Using the pattern for}%
\typeout{** the default language instead.}%
\else
\language=\csname l@#1\endcsname
\fi
#2}}
\providecommand{\BIBdecl}{\relax}
\BIBdecl

\bibitem{ScalingLaw}
J.~Kaplan, S.~McCandlish, T.~Henighan, T.~B. Brown, B.~Chess, R.~Child, S.~Gray, A.~Radford, J.~Wu, and D.~Amodei, ``Scaling laws for neural language models,'' \emph{CoRR}, vol. abs/2001.08361, 2020.

\bibitem{ScalingLaw2}
J.~Hoffmann, S.~Borgeaud, A.~Mensch, E.~Buchatskaya, T.~Cai, E.~Rutherford, D.~de~Las~Casas, L.~A. Hendricks, J.~Welbl, A.~Clark, T.~Hennigan, E.~Noland, K.~Millican, G.~van~den Driessche, B.~Damoc, A.~Guy, S.~Osindero, K.~Simonyan, E.~Elsen, J.~W. Rae, O.~Vinyals, and L.~Sifre, ``Training compute-optimal large language models,'' \emph{CoRR}, vol. abs/2203.15556, 2022.

\bibitem{DataLaw}
B.~Zhang, Z.~Liu, C.~Cherry, and O.~Firat, ``When scaling meets {LLM} finetuning: The effect of data, model and finetuning method,'' in \emph{ICLR}.\hskip 1em plus 0.5em minus 0.4em\relax OpenReview.net, 2024.

\bibitem{PartialImage}
T.~Durand, N.~Mehrasa, and G.~Mori, ``Learning a deep convnet for multi-label classification with partial labels,'' in \emph{CVPR}.\hskip 1em plus 0.5em minus 0.4em\relax Computer Vision Foundation / {IEEE}, 2019, pp. 647--657.

\bibitem{NoisySeg}
Z.~Lu, Z.~Fu, T.~Xiang, P.~Han, L.~Wang, and X.~Gao, ``Learning from weak and noisy labels for semantic segmentation,'' \emph{{IEEE} Trans. Pattern Anal. Mach. Intell.}, vol.~39, no.~3, pp. 486--500, 2017.

\bibitem{UnlabeledDetection1}
Y.~Ren, D.~Ji, and H.~Zhang, ``Positive unlabeled learning for deceptive reviews detection,'' in \emph{EMNLP}.\hskip 1em plus 0.5em minus 0.4em\relax {ACL}, 2014, pp. 488--498.

\bibitem{UnlabeledDetection2}
T.~Hu, Q.~Guo, X.~Shen, H.~Sun, R.~Wu, and H.~Xi, ``Utilizing unlabeled data to detect electricity fraud in {AMI:} {A} semisupervised deep learning approach,'' \emph{{IEEE} Trans. Neural Networks Learn. Syst.}, vol.~30, no.~11, pp. 3287--3299, 2019.

\bibitem{MultiInsProtein}
J.-S. Wu, S.-J. Huang, and Z.-H. Zhou, ``Genome-wide protein function prediction through multi-instance multi-label learning,'' \emph{IEEE/ACM Transactions on Computational Biology and Bioinformatics}, vol.~11, no.~5, pp. 891--902, 2014.

\bibitem{PairwiseRetrieval1}
Z.~Li, C.~Guo, X.~Wang, H.~Zhang, and Y.~Wang, ``Integrating listwise ranking into pairwise-based image-text retrieval,'' \emph{Knowledge-Based Systems}, vol. 287, p. 111431, 2024.

\bibitem{PairwiseRetrieval2}
Y.~Li, N.~Yang, L.~Wang, F.~Wei, and W.~Li, ``Learning to rank in generative retrieval,'' in \emph{AAAI}.\hskip 1em plus 0.5em minus 0.4em\relax {AAAI} Press, 2024, pp. 8716--8723.

\bibitem{DataUncertainty}
C.~Beck, H.~Booth, M.~El-Assady, and M.~Butt, ``Representation problems in linguistic annotations: Ambiguity, variation, uncertainty, error and bias,'' in \emph{Proceedings of the 14th Linguistic Annotation Workshop"}.\hskip 1em plus 0.5em minus 0.4em\relax Barcelona, Spain: Association for Computational Linguistics, 2020, pp. 60--73.

\bibitem{PosUnlSurvey}
J.~Bekker and J.~Davis, ``Learning from positive and unlabeled data: a survey,'' \emph{Mach. Learn.}, vol. 109, no.~4, pp. 719--760, 2020.

\bibitem{SemiSurvey}
X.~Yang, Z.~Song, I.~King, and Z.~Xu, ``A survey on deep semi-supervised learning,'' \emph{{IEEE} Trans. Knowl. Data Eng.}, vol.~35, no.~9, pp. 8934--8954, 2023.

\bibitem{paircomp2010}
J.~F{\"{u}}rnkranz and E.~H{\"{u}}llermeier, ``Preference learning and ranking by pairwise comparison,'' in \emph{Preference Learning}.\hskip 1em plus 0.5em minus 0.4em\relax Springer, 2010, pp. 65--82.

\bibitem{PairwiseSurvey2}
E.~H{\"{u}}llermeier, J.~F{\"{u}}rnkranz, W.~Cheng, and K.~Brinker, ``Label ranking by learning pairwise preferences,'' \emph{Artif. Intell.}, vol. 172, no. 16-17, pp. 1897--1916, 2008.

\bibitem{TransMatix2017}
G.~Patrini, A.~Rozza, A.~K. Menon, R.~Nock, and L.~Qu, ``Making deep neural networks robust to label noise: {A} loss correction approach,'' in \emph{CVPR}.\hskip 1em plus 0.5em minus 0.4em\relax {IEEE} Computer Society, 2017, pp. 2233--2241.

\bibitem{UnlabeledDetection2022}
S.~Zhao, Z.~Zhang, S.~Schulter, L.~Zhao, B.~G.~V. Kumar, A.~Stathopoulos, M.~Chandraker, and D.~N. Metaxas, ``Exploiting unlabeled data with vision and language models for object detection,'' in \emph{ECCV}, ser. Lecture Notes in Computer Science, vol. 13669.\hskip 1em plus 0.5em minus 0.4em\relax Springer, 2022, pp. 159--175.

\bibitem{UnlabeledDetection2024}
P.~Guo, Y.~Song, B.~Wang, J.~Liu, and Q.~Zhang, ``Plbr: A semi-supervised document key information extraction via pseudo-labeling bias rectification,'' \emph{IEEE Transactions on Knowledge and Data Engineering}, vol.~36, no.~12, pp. 9025--9036, 2024.

\bibitem{PairwiseLLMReason2025}
L.~Yuan, G.~Cui, H.~Wang, N.~Ding, X.~Wang, B.~Shan, Z.~Liu, J.~Deng, H.~Chen, R.~Xie, Y.~Lin, Z.~Liu, B.~Zhou, H.~Peng, Z.~Liu, and M.~Sun, ``Advancing {LLM} reasoning generalists with preference trees,'' in \emph{ICLR}.\hskip 1em plus 0.5em minus 0.4em\relax OpenReview.net, 2025.

\bibitem{GWSL_NIPS_2020}
Y.~Zhang, N.~Charoenphakdee, Z.~Wu, and M.~Sugiyama, ``Learning from aggregate observations,'' in \emph{NeurIPS}, 2020.

\bibitem{GWSL_ICML_2023}
Z.~Wei, L.~Feng, B.~Han, T.~Liu, G.~Niu, X.~Zhu, and H.~T. Shen, ``A universal unbiased method for classification from aggregate observations,'' in \emph{ICML}, ser. Proceedings of Machine Learning Research, vol. 202.\hskip 1em plus 0.5em minus 0.4em\relax {PMLR}, 2023, pp. 36\,804--36\,820.

\bibitem{GWSL_TPAMI}
C.~Gong, J.~Yang, J.~You, and M.~Sugiyama, ``Centroid estimation with guaranteed efficiency: {A} general framework for weakly supervised learning,'' \emph{{IEEE} Trans. Pattern Anal. Mach. Intell.}, vol.~44, no.~6, pp. 2841--2855, 2022.

\bibitem{GWSL_ICML_2024}
H.~Chen, J.~Wang, L.~Feng, X.~Li, Y.~Wang, X.~Xie, M.~Sugiyama, R.~Singh, and B.~Raj, ``A general framework for learning from weak supervision,'' in \emph{ICML}.\hskip 1em plus 0.5em minus 0.4em\relax OpenReview.net, 2024.

\bibitem{GWSL_NIPS_2023}
V.~Shukla, Z.~Zeng, K.~Ahmed, and G.~V. den Broeck, ``A unified approach to count-based weakly supervised learning,'' in \emph{NeurIPS}, 2023.

\bibitem{noisy2015}
T.~Liu and D.~Tao, ``Classification with noisy labels by importance reweighting,'' \emph{{IEEE} Trans. Pattern Anal. Mach. Intell.}, vol.~38, no.~3, pp. 447--461, 2016.

\bibitem{noisy2020b}
H.~Wei, L.~Feng, X.~Chen, and B.~An, ``Combating noisy labels by agreement: {A} joint training method with co-regularization,'' in \emph{CVPR}.\hskip 1em plus 0.5em minus 0.4em\relax Computer Vision Foundation / {IEEE}, 2020, pp. 13\,723--13\,732.

\bibitem{noisy2020ICLR}
J.~Li, R.~Socher, and S.~C.~H. Hoi, ``Dividemix: Learning with noisy labels as semi-supervised learning,'' in \emph{ICLR}.\hskip 1em plus 0.5em minus 0.4em\relax OpenReview.net, 2020.

\bibitem{noisy_tkde2024}
J.-Y. Chen, S.-Y. Li, S.-J. Huang, S.~Chen, L.~Wang, and M.-K. Xie, ``Unm: A universal approach for noisy multi-label learning,'' \emph{IEEE Transactions on Knowledge and Data Engineering}, vol.~36, no.~9, pp. 4968--4980, 2024.

\bibitem{crowd2010}
V.~C. Raykar, S.~Yu, L.~H. Zhao, G.~H. Valadez, C.~Florin, L.~Bogoni, and L.~Moy, ``Learning from crowds,'' \emph{J. Mach. Learn. Res.}, vol.~11, pp. 1297--1322, 2010.

\bibitem{crowd2016}
Y.~Zhang, X.~Chen, D.~Zhou, and M.~I. Jordan, ``Spectral methods meet {EM:} {A} provably optimal algorithm for crowdsourcing,'' \emph{J. Mach. Learn. Res.}, vol.~17, pp. 102:1--102:44, 2016.

\bibitem{crowd2017}
R.~Meng, L.~Chen, Y.~Tong, and C.~Zhang, ``Knowledge base semantic integration using crowdsourcing,'' \emph{IEEE Transactions on Knowledge and Data Engineering}, vol.~29, no.~5, pp. 1087--1100, 2017.

\bibitem{crowd2024xia}
M.~Xia, Z.~Huang, R.~Wu, G.~Lyu, J.~Zhao, G.~Chen, and H.~Wang, ``Unbiased multi-label learning from crowdsourced annotations,'' in \emph{ICML}.\hskip 1em plus 0.5em minus 0.4em\relax OpenReview.net, 2024.

\bibitem{partial_2024}
Z.~Zhang, J.~Yao, L.~Liu, J.~Li, L.~Li, and X.~Wu, ``Partial label feature selection: An adaptive approach,'' \emph{IEEE Transactions on Knowledge and Data Engineering}, vol.~36, no.~8, pp. 4178--4191, 2024.

\bibitem{PICO+}
H.~Wang, R.~Xiao, Y.~Li, L.~Feng, G.~Niu, G.~Chen, and J.~Zhao, ``Pico+: Contrastive label disambiguation for robust partial label learning,'' \emph{{IEEE} Trans. Pattern Anal. Mach. Intell.}, vol.~46, no.~5, pp. 3183--3198, 2024.

\bibitem{partial2020NIPS}
L.~Feng, J.~Lv, B.~Han, M.~Xu, G.~Niu, X.~Geng, B.~An, and M.~Sugiyama, ``Provably consistent partial-label learning,'' in \emph{NeurIPS}, 2020.

\bibitem{partial2017}
M.-L. Zhang, F.~Yu, and C.-Z. Tang, ``Disambiguation-free partial label learning,'' \emph{IEEE Transactions on Knowledge and Data Engineering}, vol.~29, no.~10, pp. 2155--2167, 2017.

\bibitem{partial2022ICML}
D.~Wu, D.~Wang, and M.~Zhang, ``Revisiting consistency regularization for deep partial label learning,'' in \emph{ICML}, ser. Proceedings of Machine Learning Research, vol. 162.\hskip 1em plus 0.5em minus 0.4em\relax {PMLR}, 2022, pp. 24\,212--24\,225.

\bibitem{CompL2020}
L.~Feng, T.~Kaneko, B.~Han, G.~Niu, B.~An, and M.~Sugiyama, ``Learning with multiple complementary labels,'' in \emph{ICML}, ser. Proceedings of Machine Learning Research, vol. 119.\hskip 1em plus 0.5em minus 0.4em\relax {PMLR}, 2020, pp. 3072--3081.

\bibitem{mil1997}
O.~Maron and T.~Lozano{-}P{\'{e}}rez, ``A framework for multiple-instance learning,'' in \emph{NeurIPS}.\hskip 1em plus 0.5em minus 0.4em\relax The {MIT} Press, 1997, pp. 570--576.

\bibitem{mil2009}
Z.~Zhou, Y.~Sun, and Y.~Li, ``Multi-instance learning by treating instances as non-i.i.d. samples,'' in \emph{ICML}, ser. {ACM} International Conference Proceeding Series, vol. 382.\hskip 1em plus 0.5em minus 0.4em\relax {ACM}, 2009, pp. 1249--1256.

\bibitem{mil2021}
B.~Li, Y.~Li, and K.~W. Eliceiri, ``Dual-stream multiple instance learning network for whole slide image classification with self-supervised contrastive learning,'' in \emph{CVPR}.\hskip 1em plus 0.5em minus 0.4em\relax Computer Vision Foundation / {IEEE}, 2021, pp. 14\,318--14\,328.

\bibitem{mil2021_2}
X.-S. Wei, H.-J. Ye, X.~Mu, J.~Wu, C.~Shen, and Z.-H. Zhou, ``Multi-instance learning with emerging novel class,'' \emph{IEEE Transactions on Knowledge and Data Engineering}, vol.~33, no.~5, pp. 2109--2120, 2021.

\bibitem{multiins_2025}
Y.-X. Zhang, Z.~Zhou, X.~He, A.~Ranjan~Adhikary, and B.~Dutta, ``Data-driven knowledge fusion for deep multi-instance learning,'' \emph{IEEE Transactions on Neural Networks and Learning Systems}, vol.~36, no.~5, pp. 8292--8306, 2025.

\bibitem{lprop2009}
N.~Quadrianto, A.~J. Smola, T.~S. Caetano, and Q.~V. Le, ``Estimating labels from label proportions,'' \emph{J. Mach. Learn. Res.}, vol.~10, pp. 2349--2374, 2009.

\bibitem{lprop2020}
C.~Scott and J.~Zhang, ``Learning from label proportions: {A} mutual contamination framework,'' in \emph{NeurIPS}, 2020.

\bibitem{lprop2022}
J.~Zhang, Y.~Wang, and C.~Scott, ``Learning from label proportions by learning with label noise,'' in \emph{NeurIPS}, 2022.

\bibitem{paircomp2021}
L.~Feng, S.~Shu, N.~Lu, B.~Han, M.~Xu, G.~Niu, B.~An, and M.~Sugiyama, ``Pointwise binary classification with pairwise confidence comparisons,'' in \emph{ICML}, ser. Proceedings of Machine Learning Research, vol. 139.\hskip 1em plus 0.5em minus 0.4em\relax {PMLR}, 2021, pp. 3252--3262.

\bibitem{pairsim2018}
H.~Bao, G.~Niu, and M.~Sugiyama, ``Classification from pairwise similarity and unlabeled data,'' in \emph{ICML}, ser. Proceedings of Machine Learning Research, vol.~80.\hskip 1em plus 0.5em minus 0.4em\relax {PMLR}, 2018, pp. 461--470.

\bibitem{pairsim_2023}
L.~Feng, S.~Shu, Y.~Cao, L.~Tao, H.~Wei, T.~Xiang, B.~An, and G.~Niu, ``Multiple-instance learning from unlabeled bags with pairwise similarity,'' \emph{IEEE Transactions on Knowledge and Data Engineering}, vol.~35, no.~11, pp. 11\,599--11\,609, 2023.

\bibitem{simconf2021}
Y.~Cao, L.~Feng, Y.~Xu, B.~An, G.~Niu, and M.~Sugiyama, ``Learning from similarity-confidence data,'' in \emph{ICML}, ser. Proceedings of Machine Learning Research, vol. 139.\hskip 1em plus 0.5em minus 0.4em\relax {PMLR}, 2021, pp. 1272--1282.

\bibitem{confdiff2023}
W.~Wang, L.~Feng, Y.~Jiang, G.~Niu, M.~Zhang, and M.~Sugiyama, ``Binary classification with confidence difference,'' in \emph{NeurIPS}, 2023.

\bibitem{posunl2008}
C.~Elkan and K.~Noto, ``Learning classifiers from only positive and unlabeled data,'' in \emph{KDD}.\hskip 1em plus 0.5em minus 0.4em\relax {ACM}, 2008, pp. 213--220.

\bibitem{posunl2015}
M.~C. du~Plessis, G.~Niu, and M.~Sugiyama, ``Convex formulation for learning from positive and unlabeled data,'' in \emph{ICML}, ser. {JMLR} Workshop and Conference Proceedings, vol.~37.\hskip 1em plus 0.5em minus 0.4em\relax JMLR.org, 2015, pp. 1386--1394.

\bibitem{posunl2017}
R.~Kiryo, G.~Niu, M.~C. du~Plessis, and M.~Sugiyama, ``Positive-unlabeled learning with non-negative risk estimator,'' in \emph{NeurIPS}, 2017, pp. 1675--1685.

\bibitem{posunl2022}
C.~Gong, Q.~Wang, T.~Liu, B.~Han, J.~You, J.~Yang, and D.~Tao, ``Instance-dependent positive and unlabeled learning with labeling bias estimation,'' \emph{{IEEE} Trans. Pattern Anal. Mach. Intell.}, vol.~44, no.~8, pp. 4163--4177, 2022.

\bibitem{posunl_NIPS2014}
M.~C. du~Plessis, G.~Niu, and M.~Sugiyama, ``Analysis of learning from positive and unlabeled data,'' in \emph{NeurIPS}, vol.~27.\hskip 1em plus 0.5em minus 0.4em\relax Curran Associates, Inc., 2014.

\bibitem{semisup_mean_teacher}
A.~Tarvainen and H.~Valpola, ``Mean teachers are better role models: Weight-averaged consistency targets improve semi-supervised deep learning results,'' in \emph{ICLR}.\hskip 1em plus 0.5em minus 0.4em\relax OpenReview.net, 2017.

\bibitem{semisup_survey}
X.~Yang, Z.~Song, I.~King, and Z.~Xu, ``A survey on deep semi-supervised learning,'' \emph{IEEE Transactions on Knowledge and Data Engineering}, vol.~35, no.~9, pp. 8934--8954, 2023.

\bibitem{semisup_gnn}
Y.~Song, Y.~Gu, T.~Li, J.~Qi, Z.~Liu, C.~S. Jensen, and G.~Yu, ``Chgnn: A semi-supervised contrastive hypergraph learning network,'' \emph{IEEE Transactions on Knowledge and Data Engineering}, vol.~36, no.~9, pp. 4515--4530, 2024.

\bibitem{semisup_MixMatch}
D.~Berthelot, N.~Carlini, I.~Goodfellow, N.~Papernot, A.~Oliver, and C.~A. Raffel, ``Mixmatch: A holistic approach to semi-supervised learning,'' in \emph{Advances in Neural Information Processing Systems}, vol.~32.\hskip 1em plus 0.5em minus 0.4em\relax Curran Associates, Inc., 2019.

\bibitem{unlunl2019}
M.~Kato, T.~Teshima, and J.~Honda, ``Learning from positive and unlabeled data with a selection bias,'' in \emph{ICLR}.\hskip 1em plus 0.5em minus 0.4em\relax OpenReview.net, 2019.

\bibitem{simunl2021}
T.~Shimada, H.~Bao, I.~Sato, and M.~Sugiyama, ``Classification from pairwise similarities/dissimilarities and unlabeled data via empirical risk minimization,'' \emph{Neural Comput.}, vol.~33, no.~5, pp. 1234--1268, 2021.

\bibitem{risk2022sugi}
M.~Sugiyama, H.~Bao, T.~Ishida, N.~Lu, and T.~Sakai, \emph{Machine learning from weak supervision: An empirical risk minimization approach}.\hskip 1em plus 0.5em minus 0.4em\relax MIT Press, 2022.

\bibitem{GBP1}
J.~S. Yedidia, W.~T. Freeman, and Y.~Weiss, \emph{Understanding belief propagation and its generalizations}.\hskip 1em plus 0.5em minus 0.4em\relax San Francisco, CA, USA: Morgan Kaufmann Publishers Inc., 2003, no. 236--239.

\bibitem{GBP2}
------, ``Generalized belief propagation,'' in \emph{NeurIPS}.\hskip 1em plus 0.5em minus 0.4em\relax {MIT} Press, 2000, pp. 689--695.

\bibitem{GBP3}
------, ``Constructing free-energy approximations and generalized belief propagation algorithms,'' \emph{{IEEE} Trans. Inf. Theory}, vol.~51, no.~7, pp. 2282--2312, 2005.

\bibitem{crowd2025}
Z.~Wang, M.~Xia, X.~Ren, J.~Zhou, G.~Lyu, T.~Hu, and H.~Wang, ``Multi-instance multi-label classification from crowdsourced labels,'' in \emph{AAAI}, vol.~39, no.~20, 2025, pp. 21\,438--21\,446.

\bibitem{factorgraph}
F.~Kschischang, B.~Frey, and H.-A. Loeliger, ``Factor graphs and the sum-product algorithm,'' \emph{IEEE Transactions on Information Theory}, vol.~47, no.~2, pp. 498--519, 2001.

\bibitem{TransMarix2022}
S.~Li, X.~Xia, H.~Zhang, Y.~Zhan, S.~Ge, and T.~Liu, ``Estimating noise transition matrix with label correlations for noisy multi-label learning,'' in \emph{NeurIPS}.\hskip 1em plus 0.5em minus 0.4em\relax Curran Associates, Inc., 2022, pp. 24\,184--24\,198.

\bibitem{TransMatrix2023}
Y.~Duan, Z.~Zhao, L.~Qi, L.~Zhou, L.~Wang, and Y.~Shi, ``Towards semi-supervised learning with non-random missing labels,'' in \emph{ICCV}, 2023, pp. 16\,121--16\,131.

\bibitem{transformer}
A.~Vaswani, N.~Shazeer, N.~Parmar, J.~Uszkoreit, L.~Jones, A.~N. Gomez, L.~Kaiser, and I.~Polosukhin, ``Attention is all you need,'' in \emph{NeurIPS}, 2017, pp. 5998--6008.

\bibitem{gumbel_softmax_E}
E.~Jang, S.~Gu, and B.~Poole, ``Categorical reparameterization with gumbel-softmax,'' in \emph{ICLR}.\hskip 1em plus 0.5em minus 0.4em\relax OpenReview.net, 2017.

\bibitem{EM1}
A.~P. Dempster, N.~M. Laird, and D.~B. Rubin, ``Maximum likelihood from incomplete data via the em algorithm,'' \emph{Journal of the royal statistical society: series B (methodological)}, vol.~39, no.~1, pp. 1--22, 1977.

\bibitem{EM2}
C.~J. Wu, ``On the convergence properties of the em algorithm,'' \emph{The Annals of statistics}, pp. 95--103, 1983.

\bibitem{CIFAR_10_100_data}
A.~Krizhevsky, G.~Hinton \emph{et~al.}, ``Learning multiple layers of features from tiny images,'' 2009.

\bibitem{STL_10_data}
A.~Coates, A.~Ng, and H.~Lee, ``An analysis of single-layer networks in unsupervised feature learning,'' in \emph{AISTATS}, 2011, pp. 215--223.

\bibitem{MNIST_data}
L.~Deng, ``The mnist database of handwritten digit images for machine learning research,'' \emph{IEEE Signal Processing Magazine}, vol.~29, no.~6, pp. 141--142, 2012.

\bibitem{F_MNIST_data}
H.~Xiao, K.~Rasul, and R.~Vollgraf. (2017) Fashion-mnist: a novel image dataset for benchmarking machine learning algorithms.

\bibitem{LLP_VAT}
K.~Tsai and H.~Lin, ``Learning from label proportions with consistency regularization,'' in \emph{ACML}, ser. Proceedings of Machine Learning Research, vol. 129.\hskip 1em plus 0.5em minus 0.4em\relax {PMLR}, 2020, pp. 513--528.

\bibitem{rank_pruning}
C.~G. Northcutt, T.~Wu, and I.~L. Chuang, ``Learning with confident examples: Rank pruning for robust classification with noisy labels,'' in \emph{UAI}.\hskip 1em plus 0.5em minus 0.4em\relax {AUAI} Press, 2017.

\bibitem{CLIP2021}
A.~Radford, J.~W. Kim, C.~Hallacy, A.~Ramesh, G.~Goh, S.~Agarwal, G.~Sastry, A.~Askell, P.~Mishkin, J.~Clark, G.~Krueger, and I.~Sutskever, ``Learning transferable visual models from natural language supervision,'' in \emph{ICML}, ser. Proceedings of Machine Learning Research, vol. 139.\hskip 1em plus 0.5em minus 0.4em\relax {PMLR}, 2021, pp. 8748--8763.

\bibitem{kuhn1955hungarian}
H.~W. Kuhn, ``The hungarian method for the assignment problem,'' \emph{Naval research logistics quarterly}, vol.~2, no. 1-2, pp. 83--97, 1955.

\bibitem{PartialLv}
J.~Lv, M.~Xu, L.~Feng, G.~Niu, X.~Geng, and M.~Sugiyama, ``Progressive identification of true labels for partial-label learning,'' in \emph{ICML}, ser. Proceedings of Machine Learning Research, vol. 119.\hskip 1em plus 0.5em minus 0.4em\relax {PMLR}, 2020, pp. 6500--6510.

\bibitem{noisy2022ICML}
S.~Liu, Z.~Zhu, Q.~Qu, and C.~You, ``Robust training under label noise by over-parameterization,'' in \emph{ICML}, ser. Proceedings of Machine Learning Research, vol. 162.\hskip 1em plus 0.5em minus 0.4em\relax {PMLR}, 2022, pp. 14\,153--14\,172.

\bibitem{PICO}
H.~Wang, R.~Xiao, Y.~Li, L.~Feng, G.~Niu, G.~Chen, and J.~Zhao, ``Pico: Contrastive label disambiguation for partial label learning,'' in \emph{ICLR}.\hskip 1em plus 0.5em minus 0.4em\relax OpenReview.net, 2022.

\bibitem{noisy2020NIPS}
S.~Liu, J.~Niles{-}Weed, N.~Razavian, and C.~Fernandez{-}Granda, ``Early-learning regularization prevents memorization of noisy labels,'' in \emph{NeurIPS}, 2020.

\bibitem{CVIR}
S.~Garg, Y.~Wu, A.~J. Smola, S.~Balakrishnan, and Z.~C. Lipton, ``Mixture proportion estimation and {PU} learning: {A} modern approach,'' in \emph{NeurIPS}, 2021, pp. 8532--8544.

\bibitem{DISTPU}
Y.~Zhao, Q.~Xu, Y.~Jiang, P.~Wen, and Q.~Huang, ``Dist-pu: Positive-unlabeled learning from a label distribution perspective,'' in \emph{CVPR}.\hskip 1em plus 0.5em minus 0.4em\relax {IEEE}, 2022, pp. 14\,441--14\,450.

\bibitem{VarPU}
H.~Chen, F.~Liu, Y.~Wang, L.~Zhao, and H.~Wu, ``A variational approach for learning from positive and unlabeled data,'' in \emph{NeurIPS}, 2020.

\bibitem{semisup_pseudo_label}
D.-H. Lee \emph{et~al.}, ``Pseudo-label: The simple and efficient semi-supervised learning method for deep neural networks,'' in \emph{Workshop on challenges in representation learning, ICML}, vol.~3, no.~2, 2013, p. 896.

\bibitem{semisup_VAT}
T.~Miyato, S.~Maeda, M.~Koyama, and S.~Ishii, ``Virtual adversarial training: {A} regularization method for supervised and semi-supervised learning,'' \emph{{IEEE} Trans. Pattern Anal. Mach. Intell.}, vol.~41, no.~8, pp. 1979--1993, 2019.

\end{thebibliography}


 

\vspace{-33pt}
\begin{IEEEbiography}
[{\includegraphics[width=1in,height=1.25in,clip,keepaspectratio]{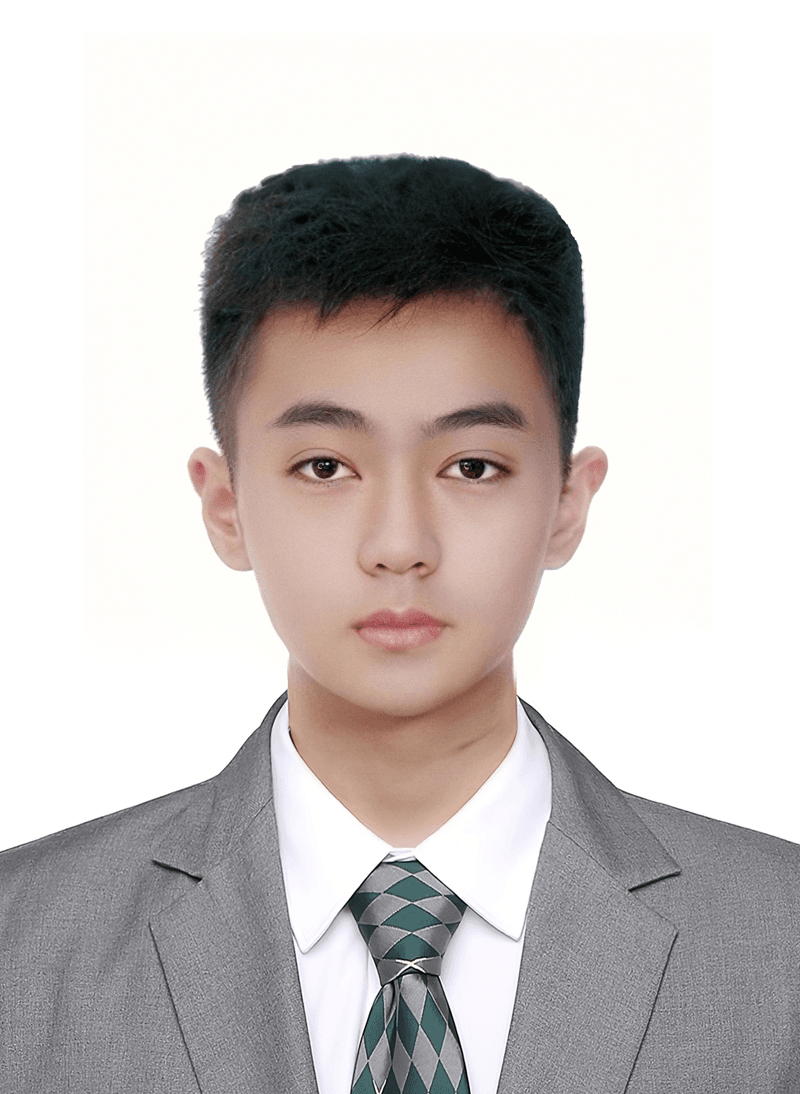}}]{Ziquan Wang}
Wang Ziquan received his bachelor's degree from Beijing University of Aeronautics and Astronautics (BUAA), in 2024. He is currently working toward his master's degree at Zhejiang University. He has published papers as the first author, including AAAI. His current research interest lies in machine learning, particularly in using data-centric approaches to achieve data-efficient and reliable fundamental models.
\end{IEEEbiography}

\vspace*{-3.2\baselineskip}
\begin{IEEEbiography}
[{\includegraphics[width=1in,height=1.25in,clip,keepaspectratio]{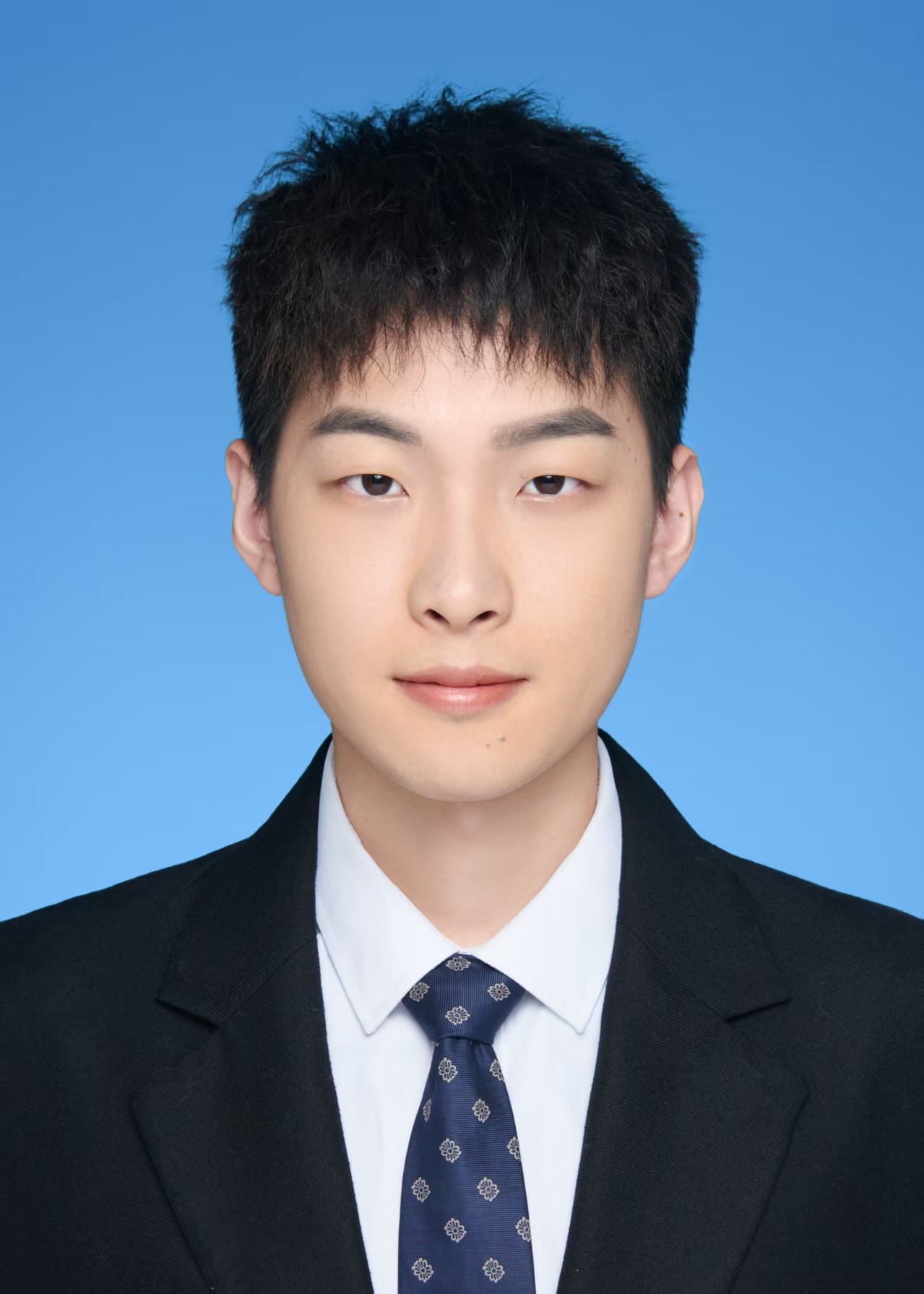}}]{Haobo Wang}
received the BEng and PhD degrees from Zhejiang University, China, in 2018 and 2023. He is currently an assistant professor in the School of Software and Technology, Zhejiang University. His paper PiCO received the Outstanding Paper Award honorable mention in ICLR 2022. He has published more than 30 papers showcased at top conferences and journals, such as TPAMI, ICLR, NeurIPS, AAAI, IJCAI, and EMNLP. His research interests include machine learning and data mining, especially on weakly-supervised learning and large language models.
\end{IEEEbiography}

\vspace*{-3.2\baselineskip}
\begin{IEEEbiography}
[{\includegraphics[width=1in,height=1.25in,clip,keepaspectratio]{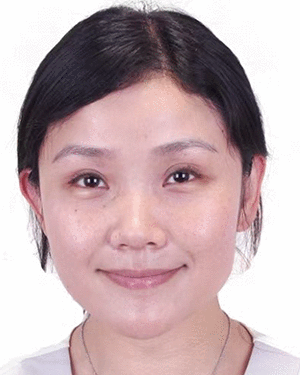}}]{Ke Chen}
received the PhD degree in computer science from Zhejiang University. She is an associate
professor with the College of Computer Science,
Zhejiang University. Her research interests include
database, large-scale data management technologies
and data privacy protection
\end{IEEEbiography}

\vspace*{-1.32\baselineskip}
\begin{IEEEbiography}
[{\includegraphics[width=1in,height=1.25in,clip,keepaspectratio]{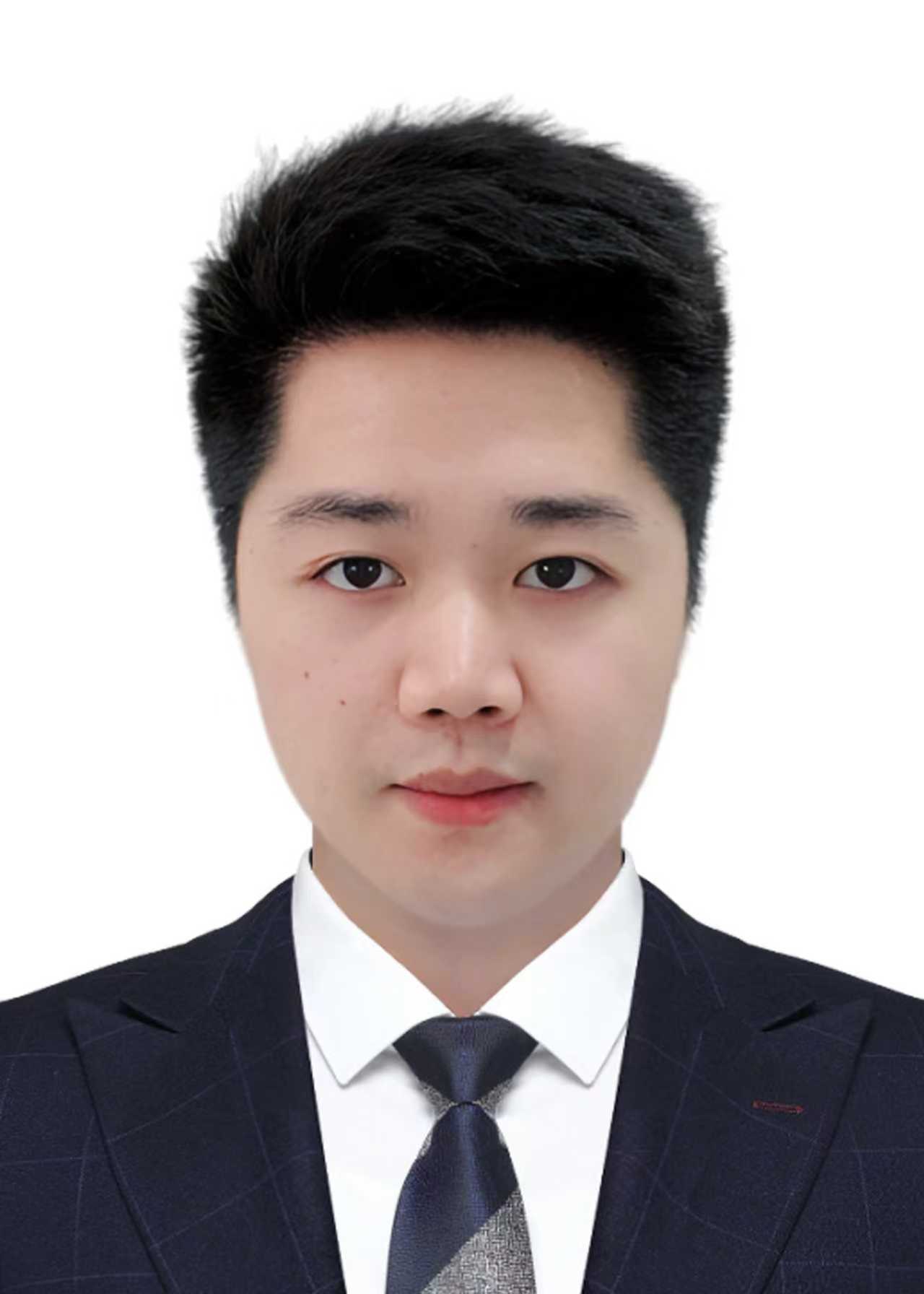}}]{Lei Feng}
received his Ph.D. degree from Nanyang Technological University, Singapore, in 2021. He is currently a Full Professor at the School of Computer Science and Engineering, Southeast University, China. His main research interests include trustworthy deep learning and AI agents. He has published more than 100 papers at premier conferences and journals. He has regularly served as an Area Chair for ICML, NeurIPS, and ICLR, and served as an Action Editor for Neural Networks and Transactions on Machine Learning Research. He has received the ICLR 2022 Outstanding Paper Award Honorable Mention and the WAIC 2024 Youth Outstanding Paper Nomination Award. He was named to Forbes 30 Under 30 China 2021, Forbes 30 Under 30 Asia 2022, and Asia-Pacific Leaders Under 30 in 2024.
\end{IEEEbiography}

\vspace*{-35.0\baselineskip}
\begin{IEEEbiography}
[{\includegraphics[width=1in,height=1.25in,clip,keepaspectratio]{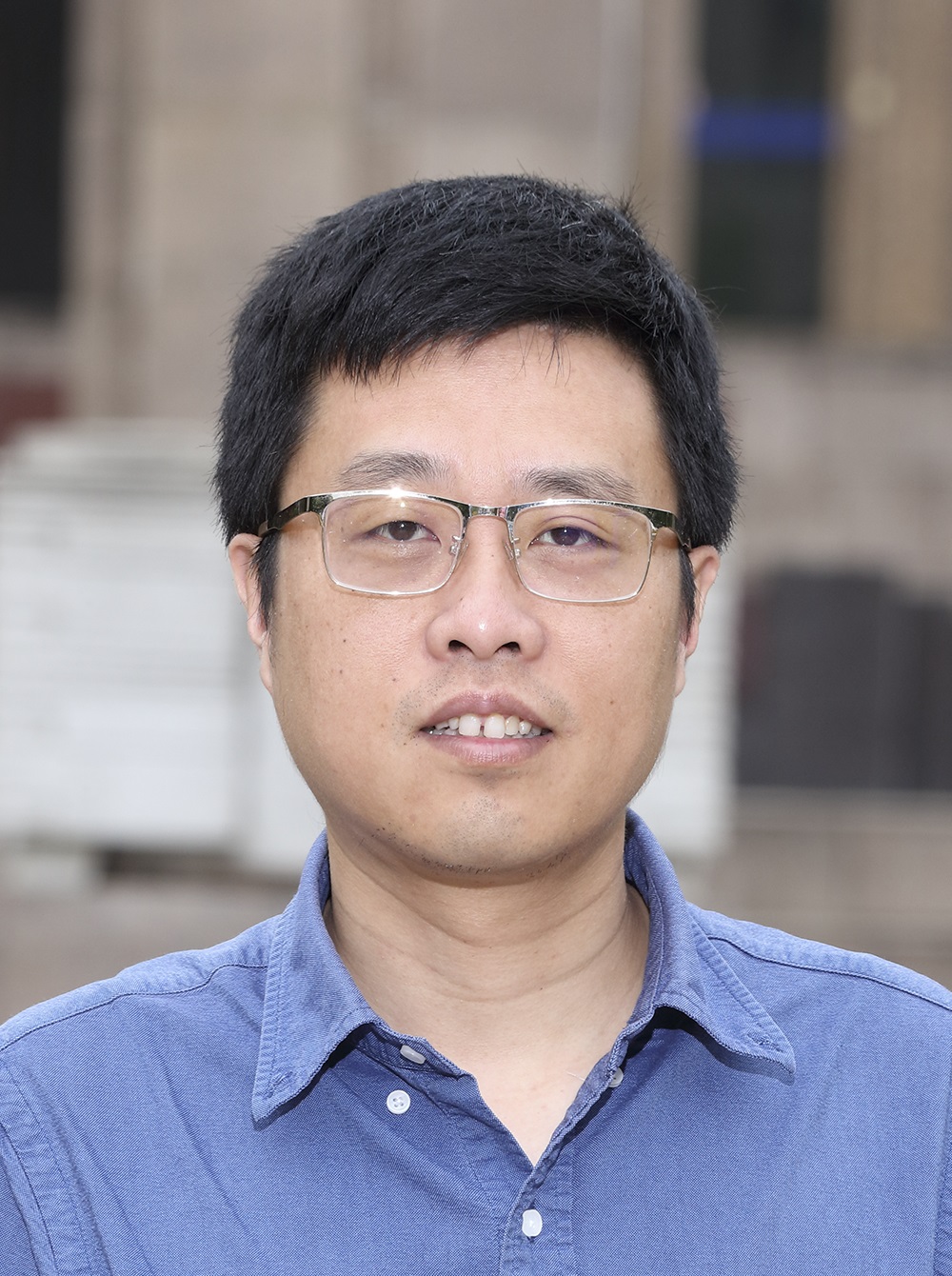}}]{Gang Chen}
(Member, IEEE) received the PhD degree in computer science from Zhejiang University, Hangzhou China. He is a professor with the College of Computer Science, Zhejiang University. He is also the director of Key Laboratory of Intelligent Computing Based Big Data of Zhejiang Province. His research interests include database management technology, intelligent computing based Big Data and massive internet systems. He is a member ACM and a standing member of the China Computer Federation Database Professional Committee.
\end{IEEEbiography}



\end{document}